\documentclass[]{template}

\usepackage[utf8]{inputenc}             
\usepackage[T1]{fontenc}                
\usepackage{url}                        
\usepackage{booktabs}                   
\usepackage{multirow}
\usepackage{colortbl}
\usepackage{multicol}
\usepackage{amsfonts}                   
\usepackage{nicefrac}                   
\usepackage{microtype}                  
\usepackage[dvipsnames]{xcolor}         

\usepackage{latexsym}

\usepackage{graphicx}
\usepackage{float}
\usepackage{subcaption}
\usepackage{wrapfig}
\usepackage{afterpage}
\usepackage{placeins}
\usepackage{lipsum}

\usepackage{bm}

\usepackage{tabularx} 
\usepackage{ragged2e} 
\newcolumntype{L}{>{\RaggedRight\hangafter=1\hangindent=0em}X}

\usepackage{enumitem}

\usepackage{amsmath}
\usepackage{amssymb}
\usepackage{mathtools}
\usepackage{amsthm}

\setboolean{logo}{true}    

\usepackage[linesnumbered,ruled,vlined]{algorithm2e}

\hypersetup{
    colorlinks=true,
    linkcolor=red,
    citecolor=Cerulean,
    filecolor=magenta,      
    urlcolor=magenta,
}

\usepackage[capitalize,noabbrev]{cleveref}
\crefname{section}{§}{§§}
\Crefname{section}{§}{§§}

\usepackage{calligra}
\DeclareMathAlphabet{\mathcalligra}{T1}{calligra}{m}{n}

\usepackage{pifont}

\theoremstyle{plain}

\theoremstyle{definition}

\theoremstyle{remark}

\renewcommand{\paragraph}[1]{\vspace{1mm}\noindent\textbf{#1}}

\DeclareCaptionLabelFormat{cont}{#1~#2\alph{ContinuedFloat}}
\usepackage[most]{tcolorbox}
\tcbset{
  promptbox/.style={
    colback=white,
    colframe=black!65,
    boxrule=0.6pt,
    arc=0pt,
    outer arc=0pt,
    left=6pt,
    right=6pt,
    top=6pt,
    bottom=6pt
  },
  taxonomygroup/.style={
    enhanced,
    colback=black!6,
    colframe=black!55,
    boxrule=0.6pt,
    arc=0pt,
    outer arc=0pt,
    left=6pt,
    right=6pt,
    top=4pt,
    bottom=4pt,
    before skip=9pt,
    after skip=4pt
  },
  taxonomyitem/.style={
    enhanced,
    colback=white,
    colframe=black!30,
    boxrule=0.4pt,
    borderline west={1.4pt}{0pt}{black!60},
    arc=0pt,
    outer arc=0pt,
    left=7pt,
    right=6pt,
    top=5pt,
    bottom=5pt,
    before skip=4pt,
    after skip=5pt
  }
}
\newtcolorbox{promptbox}[2][]{promptbox, title=#2,#1}
\tcbset{
  takeawaybox/.style={
    top=10pt,
    colback=lightgray!20,
    colframe=Black,
    colbacktitle=BurntOrange,
    enhanced,
    center,
    attach boxed title to top center={yshift=-0.1in,xshift=0.0in},
    boxed title style={boxrule=0pt,colframe=white,},
  }
}
\newtcolorbox{takeawaybox}[2][]{takeawaybox, title=#2,#1}
\tcbset{
  observationbox/.style={
    top=10pt,
    colback=lightgray!20,
    colframe=Black,
    colbacktitle=YellowGreen,
    enhanced,
    center,
    attach boxed title to top center={yshift=-0.1in,xshift=0.0in},
    boxed title style={boxrule=0pt,colframe=white,},
  }
}
\newtcolorbox{observationbox}[2][]{observationbox, title=#2,#1}

\usepackage{xspace}

\newcommand\blfootnote[1]{%
  \begingroup
  \renewcommand\thefootnote{}\footnote{#1}%
  \addtocounter{footnote}{-1}%
  \endgroup
}

\usepackage{CJK}

\title{\textsc{AdvancedMathBench}: A Benchmark Suite for Advanced Mathematical Proof Generation and Verification}

\newcommand{\paperlinks}{%
    {\normalsize\hypersetup{urlcolor=RoyalBlue}%
    \textbf{Code and data:}\enspace
    \href{https://github.com/InternLM/AdvancedMathBench}{GitHub}%
    \enspace\textbar\enspace
    \href{https://huggingface.co/datasets/debouter/AdvancedMathBench}{Dataset}%
    \enspace\textbar\enspace
    \href{https://huggingface.co/debouter/AdvancedMathBench-AutoVerifier}{AutoVerifier}}%
}

\author[1,2]{Lingkai Kong}
\author[1,3]{Zijian Wu}
\author[1,2]{Yuzhe Gu}
\author[1]{Haiteng Zhao}
\author[1,4]{Zhouqi Hua}
\author[1,2]{Wenyong Huang}
\author[1,2]{Shuang Sun}
\author[1,2]{Zhicheng Xiong}
\author[1,2]{Xiaotian Zhang}
\author[1,2]{Shuya Zhao}
\author[2]{Yan Wang}
\author[5]{Disheng Xu}
\author[1$\dagger$]{Wenwei Zhang}
\author[1$\dagger$]{Kai Chen}

\affil[1]{Shanghai AI Laboratory}
\affil[2]{Shanghai Jiao Tong University}
\affil[3]{MMLab, The Chinese University of Hong Kong}
\affil[4]{Fudan University}
\affil[5]{Great Bay University}

\begin{abstract}

Large language models (LLMs) have achieved remarkable performance on high-school and competition-level mathematics, yet their capabilities on advanced mathematics remain poorly understood. 
Existing benchmarks, however, fall short in both scope and evaluation granularity: they provide limited disciplinary coverage and often rely on final-answer correctness or coarse judgments, leaving the validity of the reasoning process inadequately assessed.
To bridge this gap, we introduce \textsc{AdvancedMathBench}, a benchmark suite designed to evaluate the reasoning capabilities of LLMs on advanced mathematical proofs. Its core generation benchmark, \textsc{ProverBench}, contains 245 problems spanning undergraduate (UG) and doctoral qualifying-exam (QE) levels.
To reliably evaluate these proofs, we develop a dedicated automatic verification pipeline that is trained on large-scale expert annotations, produces both correctness verdicts and fine-grained analyses, and exhibits strong agreement with human experts on held-out proof trajectories.
We further introduce \textsc{VerifierBench}, consisting of 888 model-generated proof trajectories paired with expert ground truth, to evaluate whether models can correctly judge proof validity and provide sound verification rationales.
Experiments show that \textsc{AdvancedMathBench} remains challenging for frontier models. On proof generation, the best-performing model, GPT-5.5-xhigh, achieves only $64.5$ and $48.9$ on the UG and QE splits, respectively. On proof verification, the best model only attains a Balanced F1 of $65.1$. Further analysis reveals a notable mismatch between proof generation and verification capabilities across models.

\end{abstract}

\begin{document}

\blfootnote{$\dagger$ Corresponding authors}

\maketitle
\par\smallskip
\noindent\paperlinks\par

\section{Introduction}

\begin{figure*}[t]
    \centering
    \includegraphics[width=0.494\textwidth]{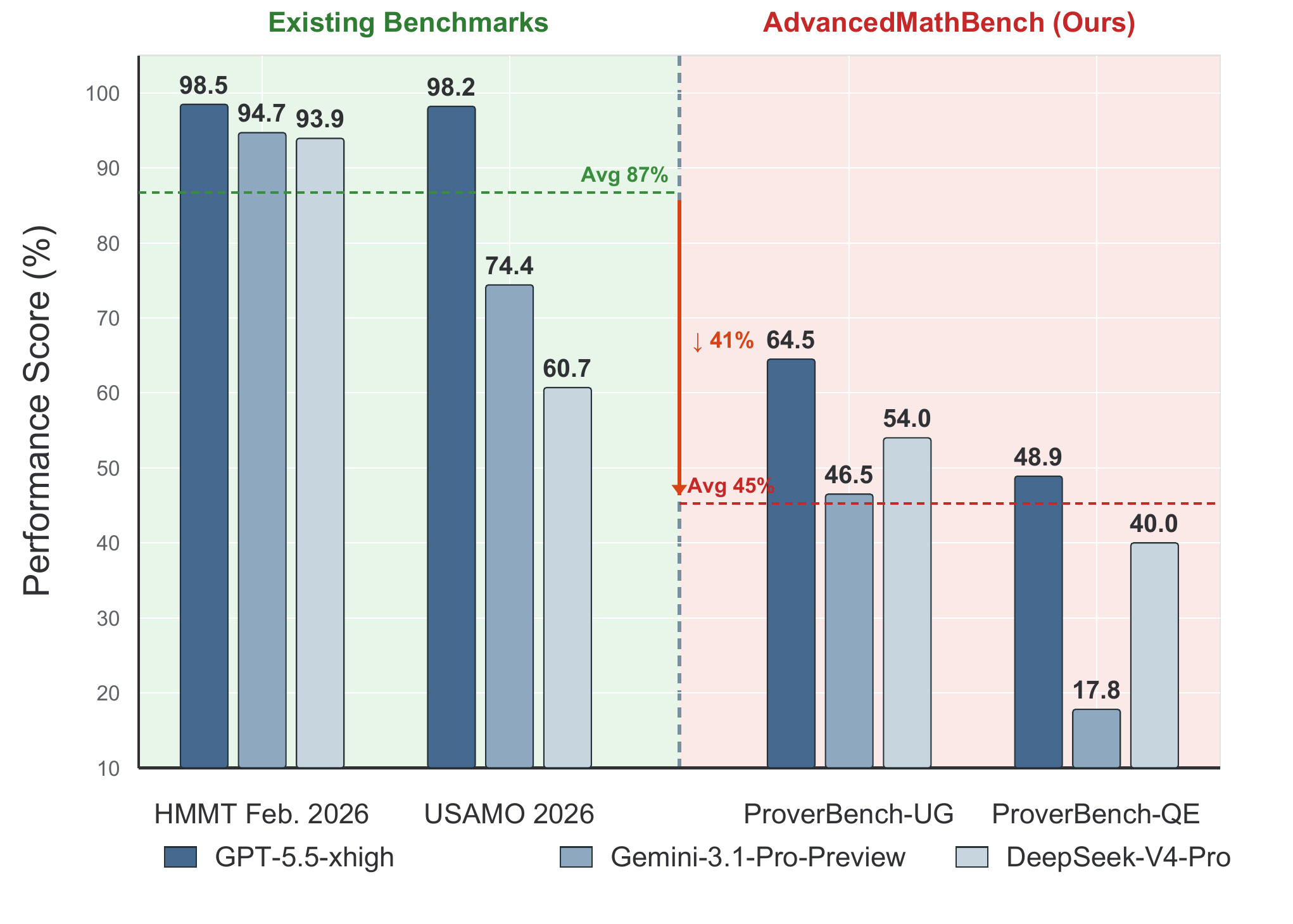}\hfill
    \includegraphics[width=0.494\textwidth]{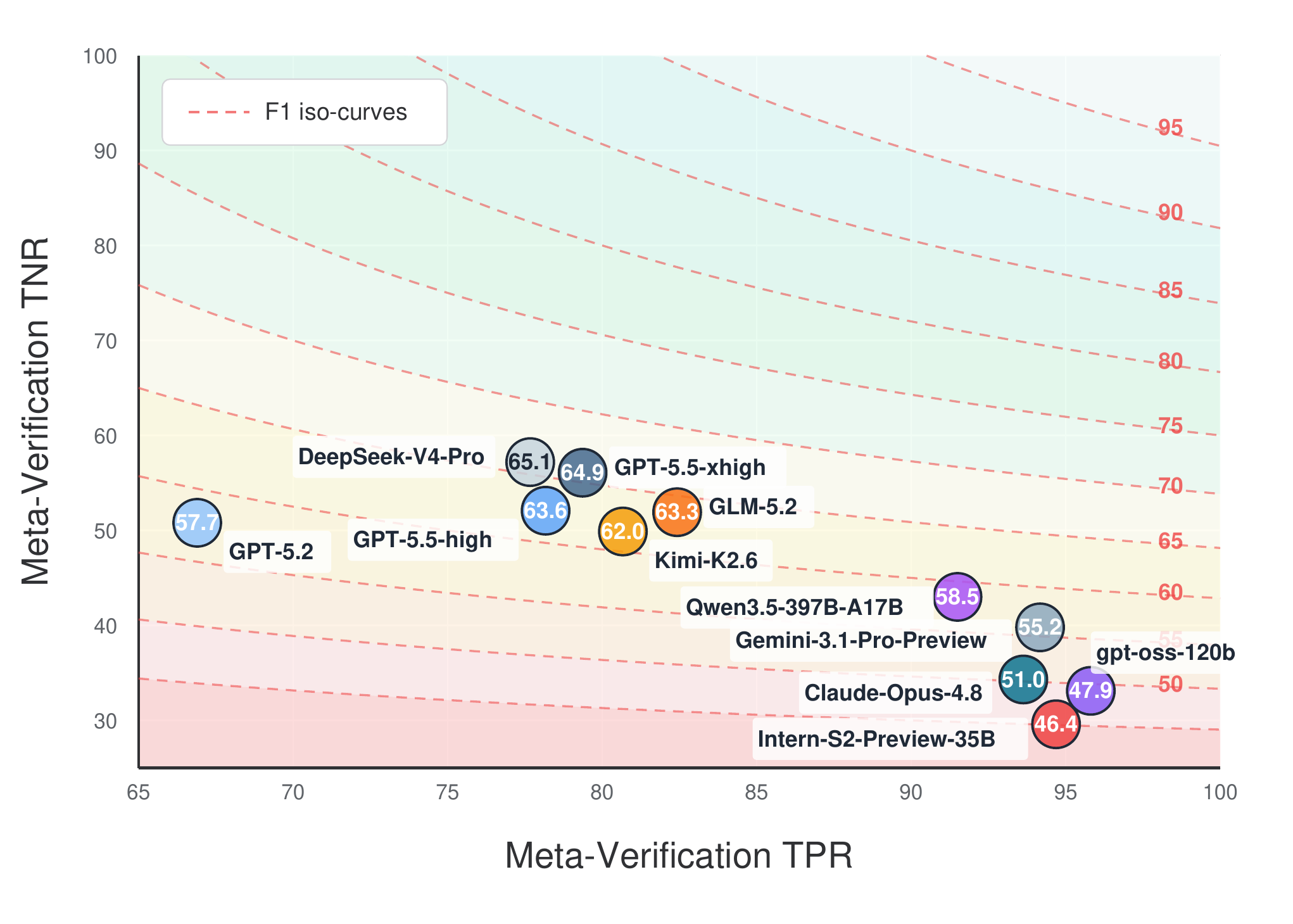}
    \caption{Model performance on \textsc{AdvancedMathBench}. Left: \textsc{ProverBench} yields substantially lower scores than HMMT and USAMO (from MathArena~\citep{dekoninck2026matharena}). Right: \textsc{VerifierBench} exposes low TNR and Balanced F1. Numbers denote Balanced F1.}
    \label{fig:teaser}
\end{figure*}

Large language models (LLMs) have achieved remarkable progress on mathematical reasoning, especially on high-school and competition-level benchmarks~\citep{cobbe2021training,hendrycks2021math,he2024olympiadbench,gao2024omnimath,bai2025interns1,zou2026interns1pro}. Yet their capabilities on advanced mathematics remain much less understood. 
Advanced mathematics requires models to construct reasoning trajectories with rigorous intermediate arguments, making mathematical proof a natural stress test of reasoning beyond final-answer prediction.

Existing benchmarks provide limited support for evaluating this capability. Most benchmarks still emphasize competition-style problems and have limited disciplinary coverage at the undergraduate, graduate, and research levels~\citep{chernyshev2024umath,fan2024hardmath,glazer2024frontiermath,dekoninck2026matharena,zhao2025interngeometry}. More importantly, many benchmarks are evaluated primarily through final-answer checking or coarse solution matching, which cannot determine whether a proof is mathematically valid.
Recent work has explored process verification and LLM-as-Judge~\citep{dekoninck2025openproof,petrov2026proofrank,lightman2023verify,zheng2024processbench,zheng2023judgingllm,luong2025robustmathreasoning,chen2025rimo},
but LLM judges still suffer from substantial bias and inconsistency.
Therefore, current benchmark methods may lead to an inaccurate estimation of the advanced mathematical capabilities of a model due to unreliable verification. 

This paper aims to provide a more rigorous foundation for measuring model behavior in advanced mathematical proof settings.
To the best of our knowledge, we present \textsc{AdvancedMathBench}, the first advanced mathematics benchmark to integrate both proof generation and verification, accompanied by a highly trustworthy verifier to ensure the rigorous evaluation of this benchmark.
It consists of two complementary components: \textsc{ProverBench}, which evaluates proof generation, and \textsc{VerifierBench}, which evaluates proof verification. 
Specifically, \textsc{ProverBench} contains 245 proof problems at undergraduate (UG) and doctoral qualifying-exam (QE) levels, covering multiple major branches of mathematics and forming a clear difficulty gradient. 
To perform reliable, fine-grained process verification on \textsc{ProverBench}, we build a dedicated automatic verification pipeline trained on large-scale expert annotations of corresponding proof trajectories.
The pipeline produces correctness verdicts, fine-grained assessments of proof quality, and error localization, enabling accurate process verification of model-generated proofs at scale. 
Complementing this generation benchmark, \textsc{VerifierBench} contains 888 model-generated proof trajectories paired with expert ground truth. In this verification task, models must judge whether a candidate proof is valid and justify the judgment. 
A meta-verifier then evaluates agreement with expert annotations in terms of validity judgments, error localization, and verification rationales.

Our experiments show that frontier LLMs still struggle with both constructing and verifying advanced mathematical proofs. 
On \textsc{ProverBench}, the state-of-the-art GPT-5.5-xhigh~\citep{openai2026gpt55} achieves only $64.5$ and $48.9$ on the UG and QE splits, respectively.
This highlights the difficulty of advanced mathematical proof generation, especially for problems requiring doctoral-level proficiency.
On \textsc{VerifierBench}, the best-performing DeepSeek-V4-Pro~\citep{deepseek2026v4} reaches a Balanced F1 of only $65.1$. Models exhibit pronounced over-acceptance and frequently miss critical errors.
Further analysis reveals that generation and verification capabilities are not aligned in either failure patterns or overall accuracy.
Stronger models produce more rigorous intermediate arguments with fewer reasoning and structural errors, yet still overlook many such errors when acting as verifiers.
Across models, stronger generators are generally better at rejecting invalid proofs, yet may penalize minor issues, thus degrading overall verification performance.
Overall, \textsc{AdvancedMathBench} provides a rigorous foundation for evaluating LLMs' capabilities in both constructing and verifying proofs in advanced mathematics.

\section{Related Work}

\paragraph{Mathematical Reasoning Benchmarks.}
Mathematical reasoning benchmarks have progressed from arithmetic and general problem solving to competition and advanced mathematics. GSM8K~\citep{cobbe2021training} and MATH~\citep{hendrycks2021math} established widely used testbeds for multi-step reasoning, while OlympiadBench~\citep{he2024olympiadbench} and Omni-MATH~\citep{gao2024omnimath} extended evaluation to harder competition problems. More recent benchmarks cover undergraduate, graduate, and research-level mathematics~\citep{chernyshev2024umath,fan2024hardmath,glazer2024frontiermath,dekoninck2026matharena}. Nevertheless, most remain confined to specific mathematical domains and problem-solving tasks evaluated through coarse final-answer checking. In contrast, \textsc{AdvancedMathBench} focuses on advanced mathematical proofs at undergraduate and doctoral levels, with broad coverage across mathematical disciplines.

\paragraph{Natural-Language Proof and Process Verification.}
Natural-language proof generation requires models to construct complete, coherent, and verifiable arguments. Prior work has progressed from constructing proof corpora and retrieval-grounded generation systems~\citep{welleck2021naturalproofs,welleck2022naturalprover} to evaluating and ranking model-generated proofs~\citep{dekoninck2025openproof,petrov2026proofrank} and, more recently, to large-scale informal theorem proving and olympiad-level proof evaluation~\citep{zhang2025deeptheorem,luong2025robustmathreasoning}. Unlike formal theorem-proving systems that provide machine-checkable guarantees~\citep{huang2018gamepad,bansal2019holist,lample2022hypertree}, natural-language proofs require fine-grained process verification to establish their validity. Process supervision and verification benchmarks therefore evaluate intermediate reasoning through step-level feedback~\citep{lightman2023verify,zheng2024processbench,wu2025opv,pandit2025hard2verify}. However, extensive studies show that general-purpose LLM-as-a-judge methods remain sensitive to prompting and task formulation, limiting their reliability across evaluation settings~\citep{zheng2023judgingllm,li2025verifybench,yosef2026rethinking,liu2026thoughtfold,hua2025imitation,zhou2025jetts}. To address this limitation, \textsc{ProverBench} employs a specialized, expert-aligned automatic verifier for rigorous process-level evaluation, while \textsc{VerifierBench} assesses the validity judgments, error localization, and rationale quality produced by LLM judges.

\begin{figure*}[t]
    \centering
    \includegraphics[width=\textwidth]{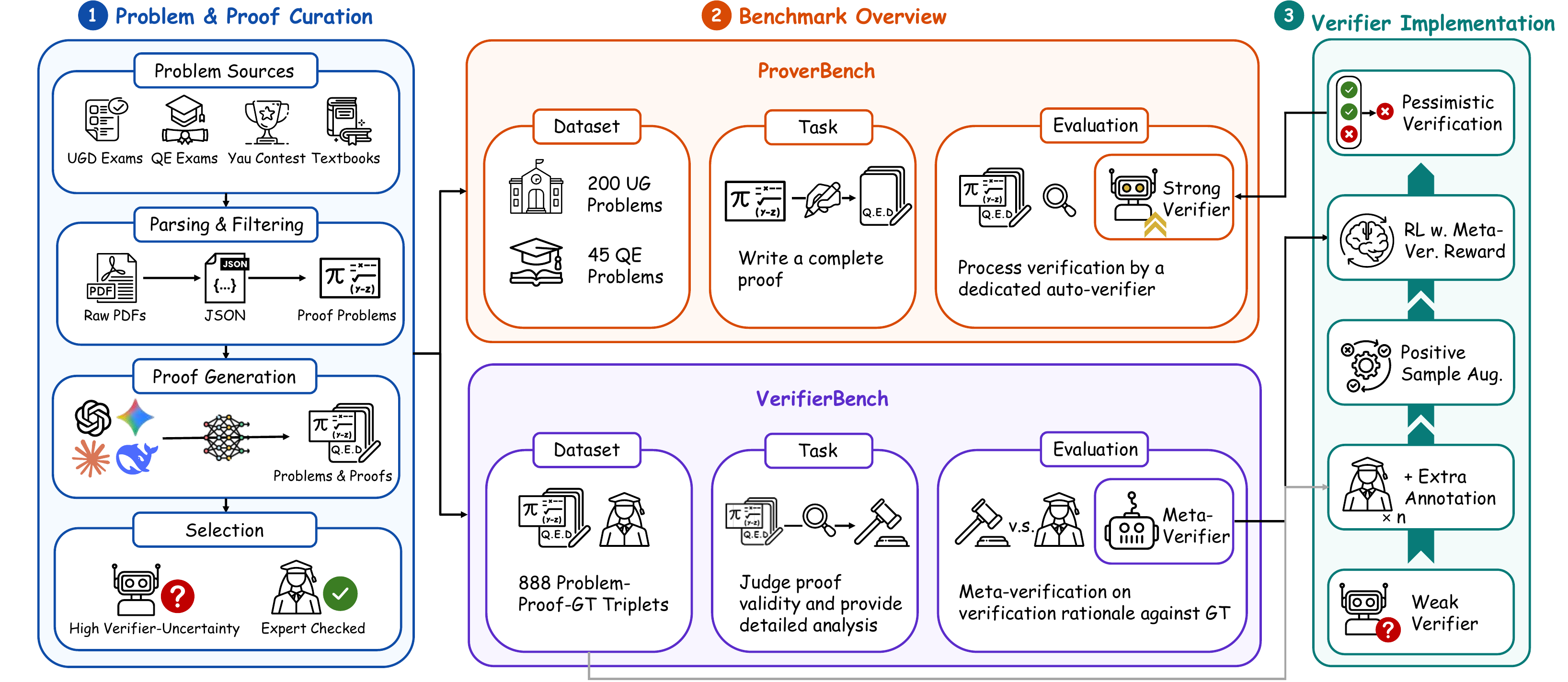}
    \caption{Overview of the \textsc{AdvancedMathBench} methodology. The figure summarizes the construction of \textsc{ProverBench} and \textsc{VerifierBench} and the implementation of our automatic verification pipeline.}
    \label{fig:methodology_schema}
\end{figure*}

\section{\textsc{AdvancedMathBench}} \label{sec:advancedmathbench}

\textsc{AdvancedMathBench} is designed to evaluate advanced mathematical proof generation and verification under rigorous process-level assessment. This section presents the benchmark construction shown in Figure~\ref{fig:methodology_schema}. Section~\ref{ppc} describes how we collect, filter, and quality-control proof problems and model-generated proof trajectories. The resulting proof-generation benchmark \textsc{ProverBench} is introduced in Section~\ref{sec:proverbench}, and the proof-verification benchmark \textsc{VerifierBench} is introduced in Section~\ref{sec:verifierbench}.

\subsection{Problem \& Proof Curation} \label{ppc}

All proof problems and model-generated proof trajectories in \textsc{AdvancedMathBench} are processed through a multi-stage curation pipeline. 
The goal is to retain problems that require substantive proof-based reasoning while removing answer-centric, noisy, or insufficiently challenging samples.

\noindent\textbf{Problem Sourcing.}
We collect problems from four complementary sources: (1) undergraduate course examinations and (2) doctoral qualifying examinations from top universities such as Stanford and Tsinghua University; (3) official problems from the S.-T. Yau College Student Mathematics Contest, covering all competition tracks; and (4) textbook exercises systematically extracted from mathematics textbooks across multiple sub-disciplines. Together, these sources provide coverage across core undergraduate mathematics, graduate-level reasoning, contest-style proof techniques, and textbook-based conceptual breadth.

\noindent\textbf{Parsing \& Filtering.}
We first parse source PDFs and extract problem statements, then filter the extracted items by task type. Answer-centric formats, such as true-or-false and fill-in-the-blank questions, are removed because they can be evaluated through answer-only guessing or final-answer matching. We retain only problems whose solutions require a rigorous proof or a complete reasoning process.

\noindent\textbf{Difficulty Pre-screening.}
We use verifier uncertainty~\citep{wu2025opv}, the entropy of repeated verification outcomes, to identify both challenging problems and plausible but error-prone proof trajectories. For each candidate problem, various top-tier LLMs are used to generate several proof trajectories. We use Intern-S2-Preview-35B~\citep{internlm2026interns2preview} to independently verify each trajectory multiple times. At the problem level, low-uncertainty problems whose generated proofs are consistently judged correct are filtered out as likely easy samples. At the proof level, trivially judged trajectories are removed, while plausible proofs that induce verifier uncertainty are retained as informative verification cases.

\noindent\textbf{Expert Quality Control.}
The remaining problems and proof trajectories are reviewed by experts with PhD-level mathematical training. 
Experts check the clarity and correctness of each problem statement, verify or revise the reference solution, and discard malformed proofs that cannot support reliable verification labels. This process yields 245 proof problems for \textsc{ProverBench} and 888 problem-proof pairs for \textsc{VerifierBench}.

\begin{figure}[!t]
    \centering
    \includegraphics[width=0.49\linewidth]{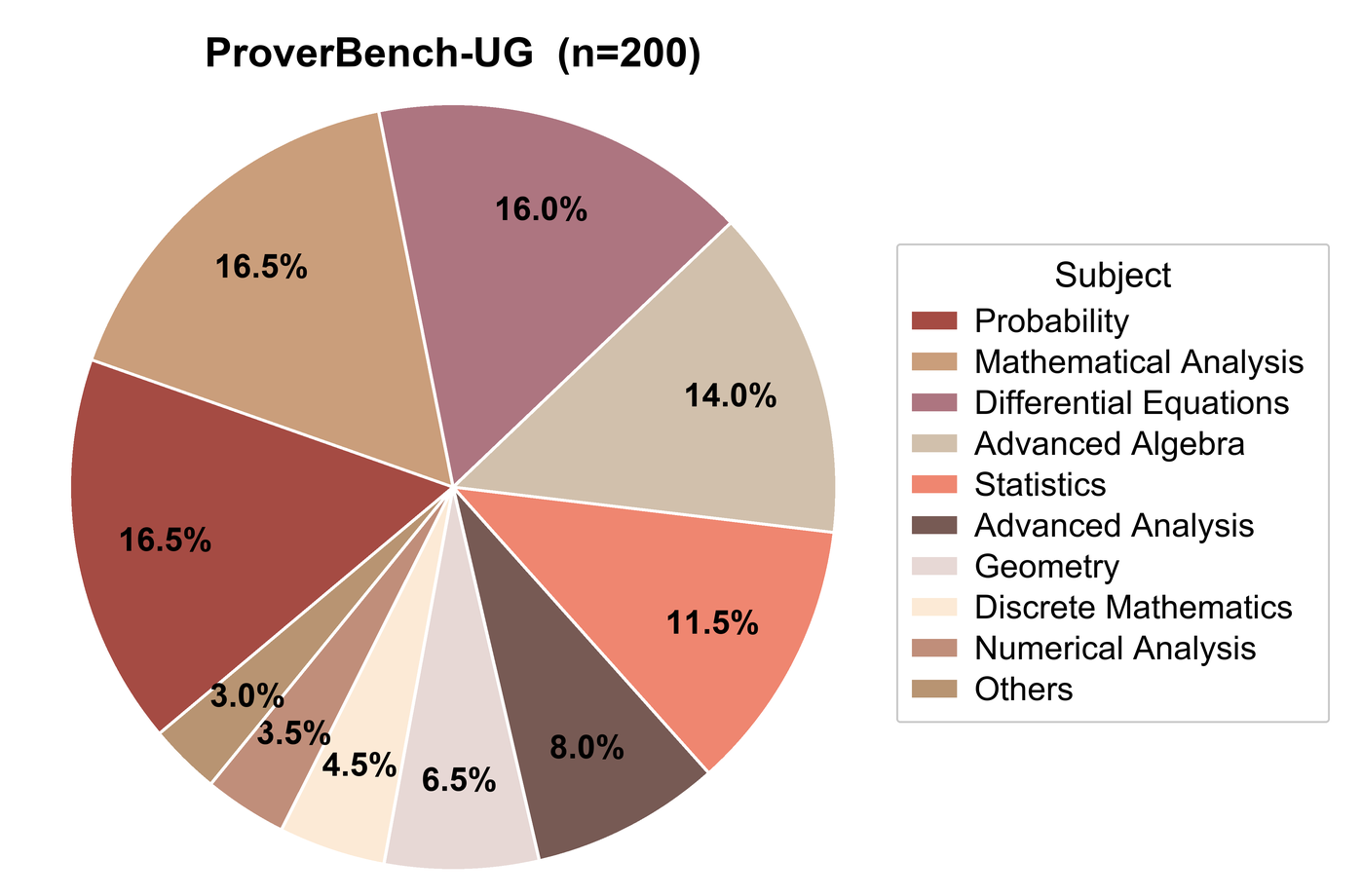}\hfill
    \includegraphics[width=0.49\linewidth]{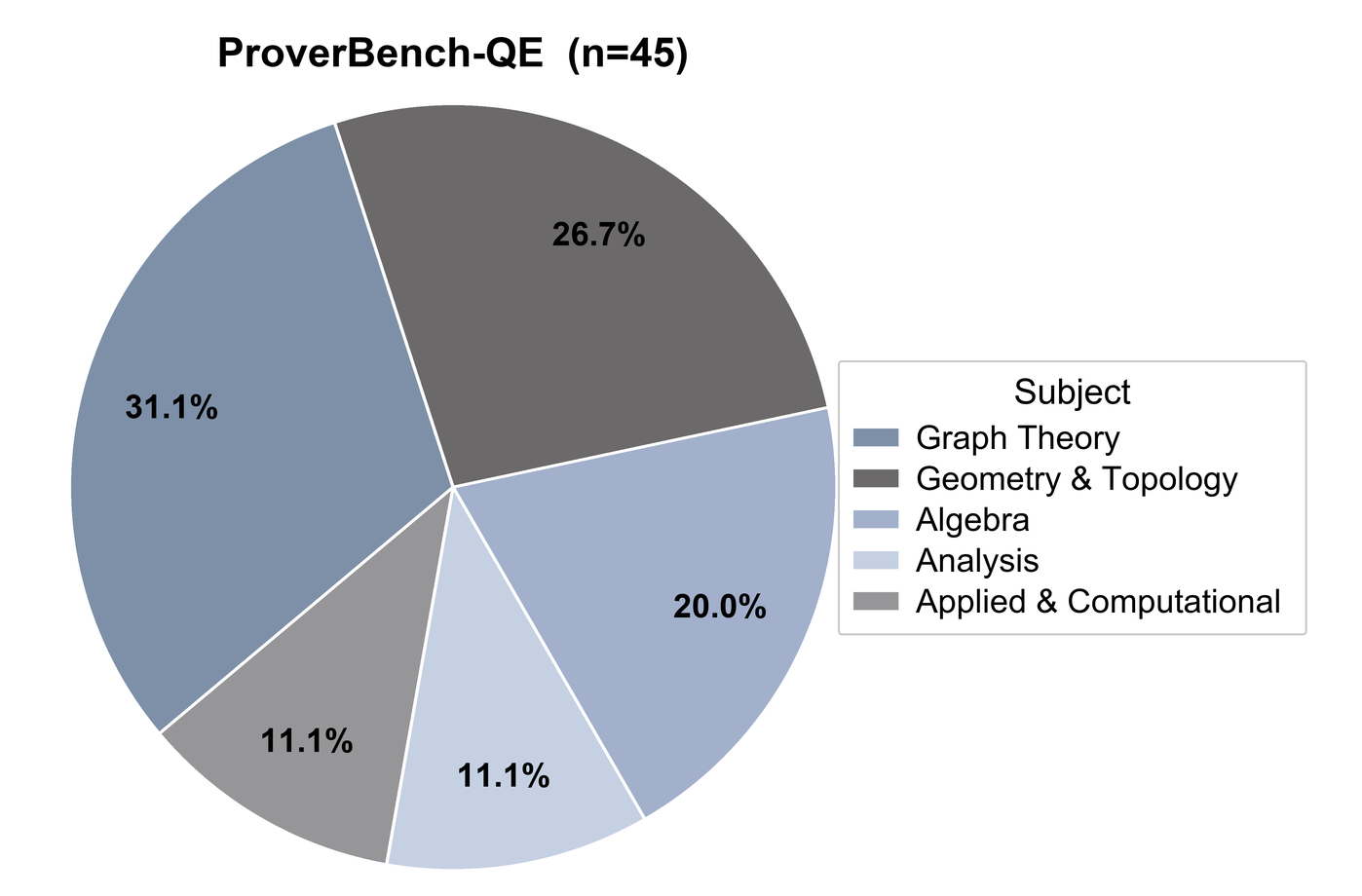}
    \caption{Subject distributions of \textsc{ProverBench}.}
    \label{fig:proverbench_distribution}
\end{figure}

\subsection{\textsc{ProverBench}} \label{sec:proverbench}

\begin{figure}[htbp]
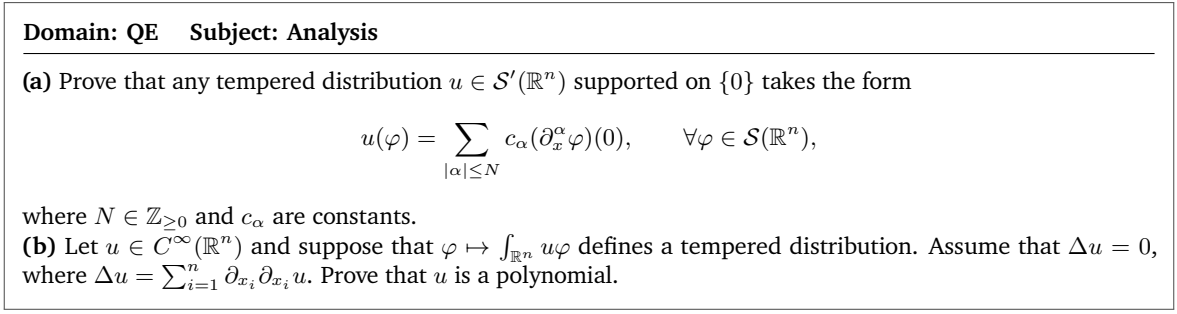

    \centering
    \setlength{\fboxsep}{7pt}
    \fcolorbox{gray!65!black}{white}{%
        \begin{minipage}{0.90\columnwidth}
        \small
        \textbf{Domain: QE \quad Subject: Analysis}
        \par\smallskip
        \hrule
        \medskip
        \textbf{(a)} Prove that any tempered distribution
        $u \in \mathcal{S}'(\mathbb{R}^{n})$ supported on $\{0\}$ takes the form
        \[
        u(\varphi)=\sum_{|\alpha|\leq N}c_{\alpha}
        (\partial_{x}^{\alpha}\varphi)(0),
        \qquad \forall\varphi\in\mathcal{S}(\mathbb{R}^{n}),
        \]
        where $N\in\mathbb{Z}_{\geq0}$ and $c_{\alpha}$ are constants.

        \textbf{(b)} Let $u\in C^{\infty}(\mathbb{R}^{n})$ and suppose that
        $\varphi\mapsto\int_{\mathbb{R}^{n}}u\varphi$ defines a tempered
        distribution. Assume that $\Delta u=0$, where
        $\Delta u=\sum_{i=1}^{n}\partial_{x_i}\partial_{x_i}u$.
        Prove that $u$ is a polynomial.
        \end{minipage}%
    }
    \caption{A representative QE-level analysis problem from \textsc{ProverBench}.}
    \label{fig:proverbench_qe_example}
\end{figure}

\textsc{ProverBench} evaluates whether models can generate complete natural-language proofs for advanced mathematical problems. The benchmark is divided into two difficulty splits according to both problem source and mathematical content. The undergraduate (UG) split contains 200 problems drawn from undergraduate course examinations and textbooks, covering core subjects such as algebra and analysis. The qualifying-exam (QE) split contains 45 problems drawn from doctoral qualifying examinations, graduate-level textbooks, and the S.-T. Yau College Student Mathematics Contest, covering more advanced subjects such as graph theory and applied and computational mathematics. The detailed subject distribution and a representative problem are shown in Figures~\ref{fig:proverbench_distribution} and~\ref{fig:proverbench_qe_example}.

Given a mathematical problem statement, a model is required to produce a step-by-step natural-language proof. 
To enable fine-grained, process-level evaluation of LLM-generated proofs, we employ an expert-aligned automatic verifier trained on large-scale expert annotations. Details of this pipeline are provided in Section~\ref{avp}.

\subsection{\textsc{VerifierBench}} \label{sec:verifierbench}

\textsc{VerifierBench} evaluates whether models can verify complete proof trajectories. For each LLM-generated proof, expert annotators perform process verification of the candidate proof, judge whether the proof is valid, provide a detailed analysis, and localize erroneous steps when the proof is invalid. This process yields 888 problem-proof-ground-truth triples.

Notably, we adopt a full-chain annotation protocol rather than first-error-only labeling~\citep{lightman2023verify,zheng2024processbench,wu2025opv}, because the earliest local issue in a model-generated proof is not always the error that determines the proof's validity. Model-generated proofs often contain minor slips before more substantive proof failures, so stopping at the first error may overemphasize trivial mistakes while missing later flaws that invalidate the global argument. Experts are therefore required to analyze the entire proof chain and distinguish between \textit{Fatal Errors}, which invalidate the global reasoning chain, and \textit{Recoverable Errors}, which are local issues that can be fixed through minor correction or additional justification, such as notation slips. 
The ground-truth annotation includes a full analysis, the step indices of recoverable errors, the index of the first fatal error, and suggested corrections.

Given a problem statement and a proof, a model is required to verify the proof process, output a validity judgment, and provide a verification rationale. We evaluate model-generated verification outputs with gpt-oss-120b~\citep{openai2025gptoss} as a meta-verifier. The meta-verifier compares each model output against the expert ground truth in terms of validity judgment, error coverage, and rationale quality, and assigns a four-level score according to the degree of alignment. Human evaluation indicates that this task is relatively straightforward given expert ground truth, with the meta-verifier achieving $96.7\%$ agreement with expert assessments.

\section{Automatic Verification Pipeline} \label{avp}

We build an expert-aligned automatic verification pipeline for \textsc{ProverBench} to evaluate model-generated proofs at scale. An overview of the implementation is shown in Figure~\ref{fig:methodology_schema}. The pipeline is designed to handle diverse proof strategies, including proofs that deviate from the reference solution, while maintaining rigorous process-level verification.

\subsection{Large-scale Annotation.}
To train the verifier on diverse and informative cases, we enlarge the proof trajectories for each problem in \textsc{ProverBench} using various additional LLMs. We also employ a simple iterative revision strategy~\citep{gao2025long} to generate higher-quality candidate proofs. The verifier-uncertainty selection strategy and expert annotation protocol described in Section~\ref{sec:advancedmathbench} are reused to annotate the most informative and challenging trajectories. Together with the original benchmark labels, this process yields approximately 2k expert annotations for verifier training.

\subsection{Positive-sample Augmentation.}
High-risk samples selected for expert annotation are naturally skewed toward negative examples because many contain proof errors. Training directly on this distribution can bias the verifier toward rejection, leading to severe false negatives and a low true positive rate. To mitigate this imbalance, we augment positive examples through proof repair. Specifically, we use expert correction suggestions to guide the original models to revise erroneous proofs, and then verify the repaired proofs with strong models such as DeepSeek-V4-Pro and GPT-5.5. Repaired proofs that pass this verification are added as positive training samples. This augmentation process yields approximately 1.2k additional positive samples.

\subsection{RL with Meta-verification Reward.}
To encourage the verifier to produce mathematically grounded analysis rather than merely predict proof polarity, we use the meta-verifier in \textsc{VerifierBench} as the judge during reinforcement learning (RL). 
The meta-verifier assigns a fine-grained reward to each verifier rollout based on its agreement with the expert ground truth. We train the verifier with GRPO~\citep{shao2024deepseekmath} starting from Intern-S2-Preview-35B.

\subsection{Pessimistic Verification.}
Errors in difficult proof problems are often plausible and intricate, and only a subset of independent verification attempts may identify the critical flaw. We therefore adopt pessimistic verification: a proof is accepted only if all independent verification passes judge it to be correct~\citep{huang2025pessimistic}. In our pipeline, the verifier performs 8 parallel verification passes for each proof and a proof is accepted only when all 8 passes accept it.

Our final verification pipeline achieves a Rough F1 of $\mathbf{82.1}$, substantially outperforming the strong GPT-5.5-xhigh baseline on a held-out evaluation set. Further evidence for its robustness, together with ablation results, is provided in Section~\ref{sec:autover_ablation}.

\section{Experiments and Results}

\subsection{Experimental Setup}

We evaluate a wide range of frontier LLMs, including (1) proprietary models: GPT-5.5-xhigh, GPT-5.5-high, GPT-5.2~\citep{openai2025gpt52}, Gemini-3.1-Pro-Preview~\citep{google2026gemini31pro}, Claude-Opus-4.8~\citep{anthropic2026claudeopus48}, and (2) open-source models: DeepSeek-V4-Pro, Qwen3.5-397B-A17B~\citep{qwen2026qwen35}, Kimi-K2.6~\citep{moonshot2026kimik26}, GLM-5.2~\citep{zhipu2026glm52}, gpt-oss-120b, and Intern-S2-Preview-35B. 
Unless otherwise specified, each model generates four independent responses with temperature $1.0$, a maximum output length of 128k tokens, and the highest reasoning effort. Reported scores are averaged across the four responses.

\subsection{\textsc{ProverBench} Results}

\textsc{ProverBench} evaluates whether models can generate complete natural-language proofs for advanced mathematical problems.
All scores are produced by our automatic verification pipeline. The \textsc{ProverBench} results are shown in the corresponding columns of Table~\ref{tab:main_results}. 

\begin{table}[t]
    \centering
    \small
    \captionsetup{font=small}
    \setlength{\tabcolsep}{5pt}
    \renewcommand{\arraystretch}{1.1}
    \begin{tabular*}{\linewidth}{@{\extracolsep{\fill}}lcccccccc}
        \toprule
        \multirow{3}{*}{Model}
        & \multicolumn{2}{c}{\textsc{ProverBench}}
        & \multicolumn{6}{c}{\textsc{VerifierBench}} \\
        \cmidrule(lr){2-3} \cmidrule(lr){4-9}
        & \multirow{2}{*}{UG} & \multirow{2}{*}{QE}
        & \multicolumn{3}{c}{Rough}
        & \multicolumn{3}{c}{Meta-Verification} \\
        \cmidrule(lr){4-6} \cmidrule(lr){7-9}
        & & & TPR & TNR & Bal. F1 & TPR & TNR & Bal. F1 \\
        \midrule
        \multicolumn{9}{c}{\textit{Proprietary Models}} \\
        \midrule
        GPT-5.5-xhigh & \textbf{64.5} & \textbf{48.9} & 78.9 & 73.3 & \textbf{76.0} & 78.9 & 55.1 & 64.9 \\
        GPT-5.5-high & 53.3 & 46.1 & 78.4 & \textbf{73.8} & \textbf{76.0} & 78.3 & 53.6 & 63.6 \\
        GPT-5.2 & 53.0 & 26.7 & 66.9 & 65.0 & 65.9 & 66.9 & 50.8 & 57.7 \\
        Gemini-3.1-Pro-Preview & 46.5 & 17.8 & 94.0 & 49.0 & 64.4 & 94.0 & 39.1 & 55.2 \\
        Claude-Opus-4.8 & 59.0 & 40.0 & \textbf{96.4} & 37.9 & 54.4 & 93.8 & 35.0 & 51.0 \\
        \midrule
        \multicolumn{9}{c}{\textit{Open-source Models}} \\
        \midrule
        DeepSeek-V4-Pro & 54.0 & 40.0 & 78.1 & 70.6 & 74.1 & 78.1 & \textbf{55.8} & \textbf{65.1} \\
        Qwen3.5-397B-A17B & 40.0 & 33.5 & 91.5 & 55.2 & 68.9 & 91.5 & 43.0 & 58.5 \\
        Kimi-K2.6 & 48.0 & 20.0 & 80.8 & 63.1 & 70.9 & 80.8 & 50.3 & 62.0 \\
        GLM-5.2 & 44.5 & 28.9 & 82.5 & 66.9 & 73.9 & 82.3 & 51.5 & 63.3 \\
        gpt-oss-120b & 20.5 & 2.2 & 95.2 & 38.7 & 55.0 & \textbf{95.3} & 32.0 & 47.9 \\
        Intern-S2-Preview-35B & 27.0 & 16.7 & 95.2 & 38.7 & 55.0 & 95.2 & 30.7 & 46.4 \\
        \bottomrule
    \end{tabular*}
    \caption{Main results on \textsc{ProverBench} and \textsc{VerifierBench}.}
    \label{tab:main_results}
\end{table}

Advanced mathematical proof generation remains challenging for frontier models. The best-performing model, GPT-5.5-xhigh, achieves only $64.5$ on the UG split and $48.9$ on the QE split. As shown in the left panel of Figure~\ref{fig:teaser}, performance on \textsc{ProverBench} is substantially lower than on the answer-centric HMMT benchmark and the competition-level proof benchmark USAMO. This gap suggests that these existing settings do not fully expose the limitations of frontier models, whereas rigorous proof generation across advanced domains provides a more discriminative evaluation.

Proprietary models outperform open-source models on average, scoring $55.3$ versus $39.0$ on UG and $35.9$ versus $23.6$ on QE. DeepSeek-V4-Pro is the strongest open-source model, achieving $54.0$ on UG and $40.0$ on QE. Although it trails GPT-5.5-xhigh by $10.5$ and $8.9$ on the two splits, respectively, it matches Claude-Opus-4.8 on QE, indicating that the gap is not uniform across models.

Doctoral problems are consistently more challenging than undergraduate ones. All evaluated models exhibit performance degradation from UG to QE, with an average drop of $17.2$.
Moreover, UG performance does not reliably predict performance on QE. For example, GPT-5.2 and Gemini-3.1-Pro-Preview achieve $53.0$ and $46.5$ on UG, respectively, but fall to $26.7$ and $17.8$ on QE, below several open-source models. These ranking changes suggest that the QE split tests advanced proof capabilities that are not adequately captured by undergraduate-level performance alone.

\subsection{\textsc{VerifierBench} Results}

\textsc{VerifierBench} evaluates whether models can reliably judge the validity of model-generated proofs and provide reasonable analysis and error localization.
We report two types of scores: (1) \textit{Rough} scores evaluate only the correctness of the binary validity judgment, while (2) \textit{Meta-Verification} scores produced by our meta-verifier additionally evaluate agreement with expert annotations in verification rationale and error localization. TPR and TNR denote accuracy on valid and invalid proofs, respectively, while Balanced F1 is their harmonic mean.
The \textsc{VerifierBench} results are shown in the corresponding columns of Table~\ref{tab:main_results}.

Overall, \textsc{VerifierBench} is challenging for both proprietary and open-source models. The best-performing model, DeepSeek-V4-Pro, achieves only $65.1$ Meta-Verification Balanced F1. The strongest proprietary and open-source models obtain similar scores, mostly between $60$ and $65$. This suggests that stronger general reasoning ability does not necessarily lead to better proof verification.

The main bottleneck lies in rejecting invalid proofs rather than accepting valid ones. As shown in the right panel of Figure~\ref{fig:teaser}, Meta-Verification TNR is consistently low, with even the best-performing model, DeepSeek-V4-Pro, reaching only $55.8$. Several models exhibit pronounced over-acceptance. gpt-oss-120b and Intern-S2-Preview-35B both obtain Meta-Verification TPR above $95$, yet their TNR scores are only $32.0$ and $30.7$, respectively. These results show that current models often fail to detect errors and logical gaps in plausible but flawed proofs.

Rough evaluation substantially overestimates verification ability. Moving from Rough to Meta-Verification reduces the average TNR by $12.3$ points and the average Balanced F1 by $9.0$ points. This gap indicates that models may produce incorrect or incomplete verification rationales while still arriving at the correct binary validity verdict. Fine-grained meta-verification is therefore necessary, as polarity matching alone cannot distinguish mathematically grounded verification from a coincidentally correct judgment.

\subsection{Error Analysis}
\label{sec:error_analysis}

We develop a hierarchical taxonomy to diagnose failure modes in advanced mathematical proof generation and verification.
On \textsc{ProverBench}, proof errors are organized into four categories: (1)~\textit{strategy-level errors}, which pursue an incorrect proof objective or misinterpret the problem; (2)~\textit{assumption errors}, which introduce unsupported premises or misapply theorems; (3)~\textit{reasoning errors}, which involve unjustified logical leaps, circular arguments, false mathematical claims, or computational mistakes; and (4)~\textit{structural errors}, which result in incomplete arguments or incorrect references to earlier steps.
On \textsc{VerifierBench}, we analyze false negatives (FNs) by categorizing why valid proofs are rejected, and false positives (FPs) by identifying which of the four proof-error categories escape detection.
We focus on GPT-5.5-xhigh and Intern-S2-Preview-35B to compare the failure modes of models at different capability levels.

\begin{table}[htbp]
    \centering
    \small
    \setlength{\tabcolsep}{6pt}
    \renewcommand{\arraystretch}{1.1}
    \begin{minipage}{0.70\linewidth}
    \centering
    \captionsetup{font=small}
    \begin{tabular*}{\linewidth}{@{\extracolsep{\fill}}lrr@{}}
        \toprule
        \raisebox{1.2ex}{Error Type} & \shortstack{GPT-5.5-\\xhigh} & \shortstack{Intern-S2-\\Preview-35B} \\
        \midrule
        Strategy-Level Errors & 41 (43.6\%) & 48 (27.1\%) \\
        Assumption Errors & 12 (12.8\%) & 17 (9.6\%) \\
        Reasoning Errors & 26 (27.7\%) & 81 (45.8\%) \\
        Structural Errors & 2 (2.1\%) & 18 (10.2\%) \\
        Other & 13 (13.8\%) & 13 (7.3\%) \\
        \midrule
        Total failures & 94 & 177 \\
        \bottomrule
    \end{tabular*}
    \caption{Distribution of proof error categories on \textsc{ProverBench}. Percentages use all failed generations as denominators.}
    \label{tab:prover_error_distribution}
    \end{minipage}
\end{table}

\begin{table}[htbp]
    \centering
    \small
    \setlength{\tabcolsep}{6pt}
    \renewcommand{\arraystretch}{1.1}
    \begin{minipage}{0.70\linewidth}
    \centering
    \captionsetup{font=small}
    \begin{tabular*}{\linewidth}{@{\extracolsep{\fill}}lrr@{}}
        \toprule
        \raisebox{1.2ex}{Error Type} & \shortstack{GPT-5.5-\\xhigh} & \shortstack{Intern-S2-\\Preview-35B} \\
        \midrule
        \textbf{False Negatives} & \textbf{95 (100.0\%)} & \textbf{17 (100.0\%)} \\
        \quad Over-strict Penalization & 94 (98.9\%) & 12 (70.6\%) \\
        \quad Interpretation Mismatch & 0 (0.0\%) & 5 (29.4\%) \\
        \quad Other & 1 (1.1\%) & 0 (0.0\%) \\
        \cmidrule(lr){1-3}
        \textbf{False Positives} & \textbf{112 (100.0\%)} & \textbf{251 (100.0\%)} \\
        \quad Strategy-Level Errors & 11 (9.8\%) & 34 (13.5\%) \\
        \quad Assumption Errors & 1 (0.9\%) & 5 (2.0\%) \\
        \quad Reasoning Errors & 74 (66.1\%) & 152 (60.6\%) \\
        \quad Structural Errors & 25 (22.3\%) & 59 (23.5\%) \\
        \quad Other & 1 (0.9\%) & 1 (0.4\%) \\
        \bottomrule
    \end{tabular*}
    \caption{Distribution of verification errors on \textsc{VerifierBench}. Percentages are computed within each model's total FNs or FPs.}
    \label{tab:verifier_error_distribution}
    \end{minipage}
\end{table}

Table~\ref{tab:prover_error_distribution} compares the proof error distributions of GPT-5.5-xhigh and Intern-S2-Preview-35B.
The stronger prover GPT-5.5-xhigh fails primarily at the strategy-selection stage. Strategy-level errors account for $43.6\%$ of GPT-5.5-xhigh's 94 failures, compared with $27.7\%$ for reasoning errors. This concentration suggests that, on difficult proof problems, even a strong model can commit to an incorrect interpretation or proof direction before substantive derivation begins. 
In contrast, Intern-S2-Preview-35B exhibits broader process-level weaknesses. Reasoning errors form its largest category ($45.8\%$), while structural errors are also considerably more frequent than for GPT-5.5-xhigh ($10.2\%$ versus $2.1\%$). Thus, the weaker prover struggles not only with strategy selection but also with sustaining rigorous intermediate reasoning and completing the resulting argument.

Table~\ref{tab:verifier_error_distribution} reports the verification error distributions.
The two verifiers exhibit sharply different FN-FP trade-offs. GPT-5.5-xhigh rejects $95$ valid proofs, with over-strict penalization accounting for $98.9\%$ of its FNs. 
It frequently treats conventional omissions, equivalent formulations, or inconsequential notation choices as substantive errors. 
Intern-S2-Preview-35B rejects only $17$ valid proofs, but accepts $251$ invalid proofs, compared with $112$ for GPT-5.5-xhigh. It therefore attains high TPR by applying a substantially more permissive verification standard.

The FPs expose a shared weakness in assessing proof reasoning and completeness. Reasoning and structural errors together account for nearly $90\%$ of the invalid proofs missed by both models. This suggests that current verifiers struggle with failures requiring global scrutiny of an argument rather than local validity checks. Notably, although GPT-5.5-xhigh produces relatively few reasoning and structural errors on \textsc{ProverBench}, it still overlooks many such errors when acting as a verifier. This asymmetry indicates that the ability to avoid an error during proof generation does not necessarily translate into the ability to detect the same error in another proof.

\subsection{Generation-Verification Relationship}
\label{sec:generation_verification_relationship}

We further explore the relationship between proof generation and verification in a matched mathematical setting. We focus on 31 QE problems shared by both \textsc{ProverBench} and \textsc{VerifierBench}, comprising 31 generation tasks and 91 verification tasks. The matched results are summarized in Table~\ref{tab:matched_generation_verification}.

\begin{table}[t]
    \centering
    \begin{minipage}{0.78\linewidth}
    \centering
    \small
    \captionsetup{font=small}
    \setlength{\tabcolsep}{5pt}
    \renewcommand{\arraystretch}{1.1}
    \definecolor{AMHeatGreen}{HTML}{DDEFE2}
    \definecolor{AMHeatLightGreen}{HTML}{EDF6E9}
    \definecolor{AMHeatYellow}{HTML}{FBF1CF}
    \definecolor{AMHeatOrange}{HTML}{FBE1C7}
    \definecolor{AMHeatRed}{HTML}{F4D6D3}
    \newcommand{\amgreen}[1]{\cellcolor{AMHeatGreen}#1}
    \newcommand{\amlightgreen}[1]{\cellcolor{AMHeatLightGreen}#1}
    \newcommand{\amyellow}[1]{\cellcolor{AMHeatYellow}#1}
    \newcommand{\amorange}[1]{\cellcolor{AMHeatOrange}#1}
    \newcommand{\amred}[1]{\cellcolor{AMHeatRed}#1}
    \begin{tabularx}{\linewidth}{p{0.40\linewidth}*{4}{>{\centering\arraybackslash}X}}
        \toprule
        \multirow{2}{*}{Model} & \multirow{2}{*}{Gen.} & \multicolumn{3}{c}{Meta-Verification} \\
        \cmidrule(lr){3-5}
        & & TPR & TNR & Bal. F1 \\
        \midrule
        GPT-5.5-xhigh & \amgreen{\textbf{62.10}} & \amred{53.26} & \amgreen{\textbf{63.69}} & \amorange{58.01} \\
        GPT-5.5-high & \amgreen{59.68} & \amred{52.64} & \amlightgreen{59.51} & \amorange{55.87} \\
        Claude-Opus-4.8 & \amlightgreen{50.81} & \amgreen{96.47} & \amorange{47.15} & \amlightgreen{63.34} \\
        DeepSeek-V4-Pro & \amlightgreen{48.66} & \amyellow{68.48} & \amgreen{62.95} & \amlightgreen{65.60} \\
        GPT-5.2 & \amyellow{37.10} & \amorange{60.87} & \amlightgreen{56.53} & \amorange{58.62} \\
        Gemini-3.1-Pro-Preview & \amyellow{37.10} & \amgreen{99.46} & \amred{41.39} & \amorange{58.45} \\
        GLM-5.2 & \amyellow{36.02} & \amlightgreen{81.52} & \amlightgreen{58.46} & \amgreen{\textbf{68.09}} \\
        Kimi-K2.6 & \amorange{34.41} & \amlightgreen{80.43} & \amyellow{49.63} & \amlightgreen{61.39} \\
        Qwen3.5-397B-A17B & \amorange{29.84} & \amgreen{92.39} & \amred{44.49} & \amyellow{60.05} \\
        Intern-S2-Preview-35B & \amred{21.77} & \amgreen{96.13} & \amred{33.68} & \amred{49.89} \\
        gpt-oss-120b & \amred{5.65} & \amgreen{95.88} & \amred{32.47} & \amred{48.52} \\
        \midrule
        \multicolumn{2}{l}{Spearman $\rho$ with Gen.} & $-0.492$ & $\mathbf{0.784}$ & $0.214$ \\
        \bottomrule
    \end{tabularx}
    \caption{Matched generation-verification results on the QE subset. Cell colors are normalized independently within each metric column, with greener cells indicating higher scores and redder cells indicating lower scores.}
    \label{tab:matched_generation_verification}
    \end{minipage}
\end{table}

Generation is only weakly associated with overall verification Balanced F1 (Spearman $\rho=0.214$). Across the first nine models in the table, generation scores decrease from $62.10$ to $29.84$, whereas Balanced F1 remains around $60$ without a consistent trend. Thus, proof-generation performance does not reliably predict overall proof-verification performance. 

The association is concentrated in rejecting invalid proofs, with generation more strongly correlated with TNR ($\rho=0.784$). Strong models such as GPT-5.5-xhigh and DeepSeek-V4-Pro achieve high scores in both generation and TNR, suggesting that detecting critical errors is more closely related to constructing rigorous proofs. However, GPT-5.5-xhigh applies an overly strict standard and rejects many valid proofs, as shown in Section~\ref{sec:error_analysis}, which lowers its TPR and overall Balanced F1. 
Conversely, weaker generators, such as Gemini-3.1-Pro-Preview and Qwen3.5-397B-A17B, may act as neglectful verifiers with inflated TPR and yield comparable Balanced F1. This highlights that TNR deserves attention in addition to the overall score when assessing verification capability.

\subsection{Ablations on Automatic Verification Pipeline} \label{sec:autover_ablation}

To better understand the contribution of each component in our automatic verification pipeline, we incrementally add components to the base verifier and compare the final pipeline with frontier LLM-as-Judge baselines. We evaluate on 94 held-out \textsc{VerifierBench} samples excluded from verifier training.

Reward design is critical to reliable verifier training. Rough-RL rewards only the binary validity judgment, whereas Meta-Ver-RL also rewards rationale quality. As shown in Table~\ref{tab:autover_ablation}, Meta-Ver-RL outperforms Rough-RL in both TNR ($72.1$ vs.\ $67.3$) and Balanced F1 ($69.0$ vs.\ $66.4$), showing that fine-grained rationale-aware supervision improves error detection even under the Rough metric.

The remaining components address the TPR-TNR trade-off. Extra annotation adds difficult, error-prone trajectories, raising TNR to $79.8$ but lowering TPR because the data skew toward invalid proofs. Positive augmentation adds verified correct proofs, restoring TPR to $81.4$ and increasing Balanced F1 to $79.1$. Finally, pessimistic verification accepts a proof only when all eight passes agree, raising TNR to $85.0$ and yielding $82.1$ Balanced F1. Together, these components improve both accuracy and class balance.

The final verifier achieves $82.1$ Rough Balanced F1, outperforming GPT-5.5-xhigh ($70.6$) and DeepSeek-V4-Pro ($69.9$). This result indicates generalization to unseen proof trajectories and supports the use of an expert-trained, domain-adapted verifier over a general-purpose LLM judge.

\begin{table}[t]
    \centering
    \begin{minipage}{0.82\linewidth}
    \centering
    \small
    \captionsetup{font=small}
    \setlength{\tabcolsep}{6pt}
    \renewcommand{\arraystretch}{1.1}
    \newcommand{\roughgain}[1]{\textcolor{ForestGreen}{\footnotesize$\blacktriangle$\,#1}}
    \newcommand{\roughdrop}[1]{\textcolor{BrickRed}{\footnotesize$\blacktriangledown$\,#1}}
    \newcommand{\roughplain}[1]{\makebox[2.2em][r]{#1}\hspace{0.3em}\makebox[3.4em][l]{}}
    \newcommand{\roughdelta}[2]{\makebox[2.2em][r]{#1}\hspace{0.3em}\makebox[3.4em][l]{#2}}
    \begin{tabular*}{\linewidth}{@{\extracolsep{\fill}}lccc}
        \toprule
        Verifier & TPR & TNR & Bal. F1 \\
        \midrule
        \multicolumn{4}{c}{\textit{Frontier-LLM-as-Judge Baselines}} \\
        \midrule
        DeepSeek-V4-Pro & \roughplain{65.0} & \roughplain{75.8} & \roughplain{69.9} \\
        GPT-5.5-xhigh & \roughplain{63.2} & \roughplain{80.0} & \roughplain{70.6} \\
        \midrule
        \multicolumn{4}{c}{\textit{Automatic Verifier Ablations}} \\
        \midrule
        Intern-S2-Preview-35B & \roughplain{76.1} & \roughplain{51.6} & \roughplain{61.5} \\
        \quad + \texttt{Rough-RL} & \roughdelta{65.5}{\roughdrop{-10.6}} & \roughdelta{67.3}{\roughgain{+15.7}} & \roughdelta{66.4}{\roughgain{+4.9}} \\
        \cmidrule(lr){1-4}
        \quad + \texttt{Meta-Ver-RL} & \roughdelta{66.1}{\roughdrop{-10.0}} & \roughdelta{72.1}{\roughgain{+20.5}} & \roughdelta{69.0}{\roughgain{+7.5}} \\
        \quad \: + \texttt{Extra-Ann.} & \roughdelta{56.8}{\roughdrop{-19.3}} & \roughdelta{79.8}{\roughgain{+28.2}} & \roughdelta{66.4}{\roughgain{+4.9}} \\
        \quad \: \: + \texttt{Pos.-Aug.} & \roughdelta{\textbf{81.4}}{\roughgain{+5.3}} & \roughdelta{77.0}{\roughgain{+25.4}} & \roughdelta{79.1}{\roughgain{+17.6}} \\
        \quad \: \: \: + \texttt{Pess.-Ver.} & \roughdelta{79.4}{\roughgain{+3.3}} & \roughdelta{\textbf{85.0}}{\roughgain{+33.4}} & \roughdelta{\textbf{82.1}}{\roughgain{+20.6}} \\
        \bottomrule
    \end{tabular*}
    \caption{Ablation of the automatic verification pipeline on the held-out set under Rough evaluation. Colored deltas denote changes relative to the base Intern-S2-Preview-35B verifier.}
    \label{tab:autover_ablation}
    \end{minipage}
\end{table}

\section{Conclusion}

We introduced \textsc{AdvancedMathBench}, a benchmark suite for process-level evaluation of advanced mathematical proof generation and verification. 
Extensive experiments show that frontier models remain far from mastering advanced proofs. They struggle to construct rigorous proofs, especially in doctoral-level mathematics, and miss critical errors during verification.
Further analysis reveals that proof generation and verification exhibit distinct, non-parallel capability patterns across models. 
We hope \textsc{AdvancedMathBench} advances LLMs toward higher-level and more rigorous mathematical reasoning, and encourages multi-dimensional, fine-grained evaluation of their capabilities.

\clearpage
\bibliographystyle{plain}
\bibliography{refs}

\clearpage
\appendix
\section{Benchmark Details}

This section provides additional details on the composition and task formats of
\textsc{AdvancedMathBench}. We first report the subject distributions of the
undergraduate (UG) and qualifying-exam (QE) splits, and then present
representative instances from \textsc{ProverBench} and \textsc{VerifierBench}.

\subsection{Subject Distributions} \label{sd}

Table~\ref{tab:proverbench_subject_distribution} reports the number and
within-split percentage of problems in each subject.

\begin{table}[htb]
    \centering
    \begin{minipage}{0.80\linewidth}
    \centering
    \small
    \renewcommand{\arraystretch}{0.92}

    \textbf{UG split ($n=200$)}
    \vspace{0.35em}

    \begin{tabular*}{\linewidth}{@{\extracolsep{\fill}}clcc}
        \toprule
        Rank & Subject & Count & Percentage \\
        \midrule
        1  & Probability & 33 & 16.5\% \\
        2  & Mathematical Analysis & 33 & 16.5\% \\
        3  & Differential Equations & 32 & 16\% \\
        4  & Advanced Algebra & 28 & 14.0\% \\
        5  & Statistics & 23 & 11.5\% \\
        6  & Advanced Analysis & 16 & 8.0\% \\
        7  & Geometry & 13 & 6.5\% \\
        8  & Discrete Mathematics & 9 & 4.5\% \\
        9 & Numerical Analysis & 7 & 3.5\% \\
        10 & Graph Theory & 3 & 1.5\% \\
        11 & Operations Research & 3 & 1.5\% \\
        \bottomrule
    \end{tabular*}

    \vspace{0.45em}

    \textbf{QE split ($n=45$)}
    \vspace{0.35em}

    \begin{tabular*}{\linewidth}{@{\extracolsep{\fill}}clcc}
        \toprule
        Rank & Subject & Count & Percentage \\
        \midrule
        1 & Graph Theory & 14 & 31.1\% \\
        2 & Geometry \& Topology & 12 & 26.7\% \\
        3 & Algebra & 9 & 20.0\% \\
        4 & Analysis & 5 & 11.1\% \\
        5 & Applied \& Computational & 5 & 11.1\% \\
        \bottomrule
    \end{tabular*}
    \caption{Subject distributions of the UG and QE splits in \textsc{ProverBench}.}
    \label{tab:proverbench_subject_distribution}
    \end{minipage}
\end{table}
\FloatBarrier

\subsection{\textsc{ProverBench} Example} \label{mpbe}

The following examples illustrate the proof generation task on the UG and QE splits, respectively. 

\begin{tcolorbox}[
    enhanced,
    colback=white,
    colframe=gray!65!black,
    boxrule=0.7pt,
    arc=0pt,
    outer arc=0pt,
    left=8pt,
    right=8pt,
    top=8pt,
    bottom=8pt
]
\small
\textbf{Domain: UG \quad Subject: Advanced Analysis}
\par\smallskip
\hrule
\medskip
On the bilateral sequence space $\ell^{2}(\mathbb{Z})$, consider the right-shift operator
\[
A(x_m)_{m\in\mathbb Z}=(x_{m-1})_{m\in\mathbb Z}.
\]
\textbf{(1)} Prove that
\[
\sigma(A)=\sigma_c(A)
=\{\lambda\in\mathbb C:|\lambda|=1\}.
\]

\textbf{(2)} Determine the distance from $A$ to the compact operators on
$\ell^{2}(\mathbb Z)$:
\[
\operatorname{dist}\!\left(A,\mathcal K(\ell^{2}(\mathbb Z))\right)
=\inf_{B\in\mathcal K(\ell^{2}(\mathbb Z))}\|A-B\|.
\]
\end{tcolorbox}

\medskip

\begin{tcolorbox}[
    enhanced,
    colback=white,
    colframe=gray!65!black,
    boxrule=0.7pt,
    arc=0pt,
    outer arc=0pt,
    left=8pt,
    right=8pt,
    top=8pt,
    bottom=8pt
]
\small
\textbf{Domain: QE \quad Subject: Algebra}
\par\smallskip
\hrule
\medskip
Let $C$ be a smooth complex projective curve of genus one, and let $p\in C$.

\textbf{(1)} Prove that there exist rational functions $f$ and $g$ on $C$
such that $f$ has a pole of order $2$ at $p$, $g$ has a pole of order $3$
at $p$, and both functions are regular on $C\setminus\{p\}$.

\textbf{(2)} Prove that $C$ is isomorphic to a smooth cubic curve in
$\mathbb{CP}^{2}$.
\end{tcolorbox}

\noindent\begin{minipage}{\columnwidth}
\captionof{figure}{Representative proof problems from \textsc{ProverBench}.}
\end{minipage}

\subsection{\textsc{VerifierBench} Example} \label{mvbe}

The example below illustrates a proof verification instance.
It contains a problem statement, a step-by-step model-generated proof,
and expert ground truth. Intermediate proof steps are omitted only for compact
presentation.

\begin{tcolorbox}[
    enhanced,
    breakable,
    colback=white,
    colframe=gray!65!black,
    colbacktitle=gray!65!black,
    coltitle=white,
    fonttitle=\bfseries,
    title={Problem Statement},
    boxrule=0.7pt,
    arc=0pt,
    outer arc=0pt,
    left=8pt,
    right=8pt,
    top=8pt,
    bottom=8pt
]
\small
Consider a smooth simple closed curve \(C\) with a bounded interior region \(\Omega\) in \(\mathbb{R}^{2}\). Formulate a variational problem to show that, among all such curves with a fixed enclosed area \(A\), the circle minimizes the perimeter \(L\).
\end{tcolorbox}

\begin{tcolorbox}[
    enhanced,
    breakable,
    colback=white,
    colframe=gray!65!black,
    colbacktitle=gray!65!black,
    coltitle=white,
    fonttitle=\bfseries,
    title={Proof Steps},
    boxrule=0.7pt,
    arc=0pt,
    outer arc=0pt,
    left=8pt,
    right=8pt,
    top=8pt,
    bottom=8pt
]
\footnotesize
\noindent\textbf{Step 0.}
Let \(C\subset \mathbb{R}^2\) be a smooth simple closed curve bounding a bounded region \(\Omega\). Choose a smooth \(2\pi\)-periodic parametrization
\[
r(t)=(x(t),y(t)),\qquad t\in[0,2\pi],
\]
with \(r(0)=r(2\pi)\), and assume \(r\) is regular, i.e. \((x'(t),y'(t))\neq (0,0)\) for all \(t\). The perimeter of \(C\) is the functional
\[
L[x,y]=\int_{0}^{2\pi}\sqrt{x'(t)^2+y'(t)^2}\,dt.
\]

\medskip
\centerline{\(\cdots\)}
\medskip

\noindent\textbf{Step 5.}
Let \(T(s)=(\dot{x}(s),\dot{y}(s))\) be the unit tangent vector. Because \(s\) is arc length,
\[
\|T(s)\|^2=\dot{x}(s)^2+\dot{y}(s)^2=1.
\]
Differentiate \(T\):
\[
\begin{aligned}
T'(s)&=(\ddot{x}(s),\ddot{y}(s))\\
&=(\lambda \dot{y}(s),-\lambda \dot{x}(s))\\
&=\lambda(\dot{y}(s),-\dot{x}(s)).
\end{aligned}
\]
Define the right unit normal vector \(N(s)\) by
\[
N(s)=(\dot{y}(s),-\dot{x}(s)).
\]
Then \(\|N(s)\|=1\) and \(N(s)\perp T(s)\). Therefore,
\[
T'(s)=\lambda N(s).
\]
By the Frenet formula in the plane, the signed curvature \(\kappa(s)\) satisfies \(T'(s)=\kappa(s)N(s)\). Comparing gives
\[
\kappa(s)\equiv \lambda,
\]
i.e. any stationary curve has constant curvature.

\medskip
\centerline{\(\cdots\)}
\medskip

\noindent\textbf{Step 9.}
Conclusion (variational formulation and minimizer): The variational problem is
\[
\min\left\{\,L[C]: C \text{ smooth simple closed},\ \operatorname{Area}(\Omega)=A_0\,\right\}.
\]
If \(C\) is a minimizer, then by the Lagrange-multiplier method it is stationary for \(J_\lambda=L+\lambda(A-A_0)\), hence has constant curvature and therefore is a circle. Since a circle with area \(A_0\) exists and has perimeter \(2\sqrt{\pi A_0}\), the minimal perimeter is achieved by that circle, giving the isoperimetric inequality
\[
\boxed{\,L^2\ge 4\pi A\,}
\]
for any smooth simple closed curve enclosing area \(A\), with equality if and only if \(C\) is a circle up to rigid motions.
\end{tcolorbox}

\begin{tcolorbox}[
    enhanced,
    breakable,
    colback=white,
    colframe=gray!65!black,
    colbacktitle=gray!65!black,
    coltitle=white,
    fonttitle=\bfseries,
    title={Expert Annotations},
    boxrule=0.7pt,
    arc=0pt,
    outer arc=0pt,
    left=8pt,
    right=8pt,
    top=8pt,
    bottom=8pt
]
\small

\textbf{Reviewer Comment:} Step 9 contains a fatal error. Without first proving the existence of a global minimizer, the proof directly treats the unique stationary point obtained from the variational calculation as a global minimum, although stationarity is only a necessary condition. Step 5 also contains a minor sign-convention issue: the proof defines the right normal vector but applies the curvature formula as if using the left normal convention.

\medskip

\textbf{First Fatal Error Step:} 9. 

\medskip

\textbf{Recoverable Error Step:} 5. 
\end{tcolorbox}

\section{Prompts}

This section provides the exact prompts used for proof generation, proof
verification, and meta-verification. 

\subsection{Proof Generation Prompt} \label{pgp}

The following prompt is used for the proof generation task in \textsc{ProverBench}.

\begin{tcblisting}{
    promptbox,
    breakable,
    listing only,
    listing options={
        basicstyle=\ttfamily\footnotesize,
        breaklines=true,
        columns=fullflexible,
        keepspaces=true,
        showstringspaces=false
    }
}
Your task is to write a proof solution to the following problem. Your proof will be graded by human judges for accuracy, thoroughness, and clarity. When you write your proof, follow these guidelines:

- You are creating a proof, not a proof outline. Each step should be carefully explained and documented. If not properly explained, the judge will assume that you cannot explain it, and therefore decrease your grade.
- You can use general theorems and lemmas, but only if they are well-known. As a rule of thumb: if the result has a name and is famous enough to have a Wikipedia page or something similar to describe it, it is allowed. Any result from papers that would not be taught in high school or low-level bachelor courses in mathematics should not be used. Any use of such results will immediately give you a zero grade.
- Do not skip computation steps in your proof. Clearly explain what transformations were done and why they are allowed in each step of a calculation.
- Your proof should be self-contained.
- If you are not sure about a specific step, or do not know how to prove an intermediate result, clearly state this. It is much preferable to indicate your uncertainty rather than making incorrect statements or claims.

### FORMATTING GUIDELINES

- Split your proof into steps, each enclosed in <step>...</step>.
- Only return the proof solution, in detailed steps, no headers, no explanations, no other text, only the <step>...</step> ... <step>...</step> tags.
- You should use correct LaTeX notation to write equations and mathematical symbols. You should encompass these equations in appropriate symbols ("\\(" and "\\)" for inline math and "\\[" and "\\]" for block math) to enhance the clarity of your proof. Do not use any unicode characters.
- If the problem has a final answer, put your final answer within \\boxed{{}}.

{problem}
\end{tcblisting}

\subsection{Proof Verification Prompt} \label{pvp}

The following prompt is used for the proof verification task in \textsc{VerifierBench} and the automatic verification pipeline in \textsc{ProverBench}. The reference solution is provided only to the automatic verification pipeline.
The required \texttt{first\_error\_step} field is used solely for a naive voting mechanism and is discarded during meta-verification.

\begin{tcblisting}{
    promptbox,
    breakable,
    listing only,
    listing options={
        basicstyle=\ttfamily\footnotesize,
        breaklines=true,
        columns=fullflexible,
        keepspaces=true,
        showstringspaces=false
    }
}
You are an **expert math proof grader**. You are judging the correctness of an LLM-generated proof for a math problem.

### Input

Your input will consist of:

* **Problem Statement**: A mathematical problem that the proof is attempting to solve.
* **Reference Solution (optional)**: When present, a correct solution or proof for reference. This is **not necessarily the only valid solution**. If the problem requires a final numeric or algebraic answer, this section contains the correct answer, which should be the only accepted final answer (though alternative reasoning paths are valid). If it is missing or empty, judge using only the problem statement and the proof.
* **Proof Solution**: The proof that you need to evaluate. This proof may contain errors, omissions, or unclear steps. The proof was generated by another language model. The proof has a clear step-wise structure: each step is wrapped as `<step idx> ... </step idx>`, where `idx` is a zero-based step index.

### Task

Analyze the proof carefully.

**Core principles (in order of precedence):**
1) **Mathematical validity** of the proof's reasoning and conclusion.
2) **Problem constraints** (e.g., unique required final value; forbidden tools if stated).
3) **Reference solution** (when present) as an anchor for sufficiency, not exclusivity.

**Alternative-approach policy:**
- If the proof uses a different but valid method, accept it as long as the reasoning is mathematically sound and satisfies the problem constraints.
- **Do not penalize** solely for re-ordering steps, using different lemmas, or giving a correct shortcut, **unless** the problem forbids it.

**Rigor and evidence:**
- Treat a claim as correct **only if it is adequately justified** within the proof (not merely asserted).
- If a step is plausible but under-justified, note the gap explicitly and judge conservatively.

**What to produce:**
- Identify logical errors, incorrect steps, or unjustified leaps.
- Give a **detailed assessment** of the proof's correctness and rigor.
- Determine whether the proof is **fully correct**, **partially correct**, or **incorrect**, and justify this judgment clearly.

### Output Format

Respond with **only** well-formed XML using the structure below. Do not include any extra text or Markdown.

**Requirements:**
- `<assessment>` must be a **detailed analysis** explaining your reasoning step-by-step. Reference specific steps (`idx`) where relevant.
- `<errors>` must be a list of specific issues (empty if the proof is fully correct).
- `<first_error_step>` must be the **index of the earliest step (`idx`) where a mathematical error or unjustified leap first occurs**.
  - If no error exists, set `<first_error_step>` to `-1`.

Example output:

<assessment>The proof shows a good understanding of the main idea, but has some unclear reasoning and minor mistakes...</assessment>
<errors>
  1. specific error 1,
  2. specific error 2,
  ...
</errors>
<first_error_step>2</first_error_step>

----------------------------------
**Problem Statement**
{problem}

**Reference Solution (optional)**
{human_solution}

**Proof Solution**
{solution}
\end{tcblisting}

\subsection{Meta-Verification Prompt} \label{mvp}

The following prompt is used both to evaluate model-generated verification outputs against expert annotations in \textsc{VerifierBench} and to provide reward signals for verifier RL. For benchmark scoring and RL reward assignment, \texttt{EXACT\_MATCH}, \texttt{BASIC\_MATCH}, \texttt{POOR\_MATCH}, and \texttt{WRONG\_POLARITY} receive scores of $1.0$, $0.75$, $0.5$, and $0$, respectively.

\begin{tcblisting}{
    promptbox,
    breakable,
    listing only,
    listing options={
        basicstyle=\ttfamily\footnotesize,
        breaklines=true,
        columns=fullflexible,
        keepspaces=true,
        showstringspaces=false
    }
}
You are an **expert meta-verifier for math-proof verification quality**.

You are given:

- a **Problem Statement**
- a **Reference Solution** (Optional)
- a **Proof Solution** (structured as `<step idx> ... </step idx>`)
- a **Solution Verification** (the verifier's XML output)
- **Human Ground Truth (GT)** labels

Your task is to evaluate the **verifier output** against the **GT oracle**.

## Core Rule

Human GT is the **oracle ground truth**. Evaluate the verifier **strictly against GT**.
Do **not** re-grade the proof independently.

Step indices refer to `<step idx> ... </step idx>`, starting from 0.

## Verifier Output Format

The `Solution Verification` is an XML output with:

- `<assessment>`: a detailed step-by-step analysis that may reference proof steps by `idx`
- `<errors>`: a list of specific issues; it is empty if the verifier judges the proof fully correct

For this task, infer the verifier's overall status primarily from `<errors>`:

- if `<errors>` is empty, treat the verifier as judging the proof **correct**
- if `<errors>` contains one or more substantive issues, treat the verifier as judging the proof **incorrect**
- use `<assessment>` to determine which claimed issues were identified, how they are justified, and whether they align with GT

## GT Fields

- `reviewer_comment`: oracle explanation of the real issue(s)

  > **Note on `reviewer_comment`**: When reading this field, disregard any
  > numeric scores (e.g., "score X", "X/7 points", "full score X"),
  > references to a scoring rubric or marking scheme (e.g., "according to the
  > rubric", "scoring point", "marking scheme"), or citations of a
  > reference/model solution used purely for scoring purposes. Focus only on
  > the substantive step-by-step verification reasoning.
  >
- `first_fatal_error_step`: first fatal error step index; `-1` means no fatal error
- `recoverable_error_steps`: recoverable error step indices before the first fatal error; `[-1]` means none

A **fatal error** is a serious error that breaks global correctness.
A **recoverable error** is a local, minor, fixable issue that does not break the whole proof, such as a typo, a minor technical slip, a small omitted detail, a local under-justification, or a small local defect.

GT says correct if and only if both `first_fatal_error_step == -1` and `recoverable_error_steps == [-1]`, otherwise incorrect.

## What to Evaluate

1. whether GT treats the proof as correct or incorrect
2. whether the verifier treats the proof as correct or incorrect
3. which verifier-identified errors actually match GT
4. whether the verifier found the GT first fatal error, if any
5. whether the verifier found all relevant GT recoverable errors
6. whether the verifier introduced false positives
7. the verifier's overall correctness and completeness relative to GT

## Matching Rule

A verifier-identified issue counts as a correct GT match only if:

1. **Step match**: it points to the same GT-labeled erroneous step
2. **Reason match**: its reason is materially aligned with `reviewer_comment`
3. **Grounding**: the claim is supported by the Proof Solution text

Exact wording is not required, but the underlying issue must materially match GT.

## False Positives and Misses

A verifier claim is a **false positive** if:

- it flags a step not labeled erroneous by GT, or
- its reason does not materially align with GT, or
- the claim is not supported by the Proof Solution text

A GT issue is **missed** if:

- the verifier does not identify it, or
- points to the wrong step, or
- gives a materially wrong reason

## First Fatal Error

If `first_fatal_error_step != -1`, judge whether the verifier correctly identifies the GT first fatal error under the matching rule above.

If `first_fatal_error_step == -1`, treat first-fatal matching as not applicable.

The verifier does **not** need to explicitly call the error "fatal".

## Completeness Cutoff

- If a fatal error exists, only GT recoverable errors **before** the first fatal error matter for completeness evaluation
- Errors after the first fatal error are irrelevant for completeness evaluation
- If the verifier also raises issues **after** the first fatal error, do not count them as false positives merely because GT does not annotate them
- However, post-fatal claims that are unsupported by the Proof Solution text may still count as false positives
- If no fatal error exists, evaluate completeness over all GT recoverable errors

## Feedback Levels

Assign exactly one:

### EXACT_MATCH

Use when:

- GT says correct and verifier also says correct, **or**
- GT says incorrect and the verifier finds **all GT-relevant errors that matter**, including:
  - the first fatal error if one exists
  - all relevant recoverable errors under the cutoff rule
- the verifier's reasons are materially aligned with GT
- there are **no material false positives**

Notes:

- Any **material** false positive disqualifies `EXACT_MATCH`.
- Minor wording differences are acceptable if the underlying issue still matches GT.

### BASIC_MATCH

Use when:

- GT says incorrect
- verifier also says incorrect
- and the verifier's overall judgment is **basically aligned** with GT

More specifically:

- if GT has a fatal error, the verifier correctly finds the **first fatal error**
- but the verifier may miss some relevant GT recoverable errors
- and/or introduce **limited false positives** that do not overturn the main GT alignment

If GT has **no fatal error**, use this level when:

- the verifier correctly finds **some but not all** GT recoverable errors,
- and the overall match to GT is still reasonably solid

Notes on false positives:

- `BASIC_MATCH` can still apply if false positives are **minor, limited, and secondary**
- if false positives are numerous, central, or substantially distort the verifier's reasoning, do **not** use `BASIC_MATCH`

### POOR_MATCH

Use when:

- GT says incorrect
- verifier also says incorrect
- but the verifier's match to GT is **weak, incomplete, or poorly aligned**

This includes cases where:

- if GT has a fatal error, the verifier does **not** correctly find the **first fatal error**
- it only finds recoverable issues, later issues, or weakly related issues instead of the true key issue
- its matching to GT is substantially incomplete or poorly grounded
- it introduces **material or numerous false positives**
- the false positives partially replace, obscure, or distort the real GT issues

If GT has **no fatal error**, use this level when:

- the verifier's matching to GT recoverable errors is weak,
- or it is heavily mixed with false positives,
- or the overall alignment with GT is not reliable enough for `BASIC_MATCH`

### WRONG_POLARITY

Use when:

- GT says correct but verifier says incorrect, **or**
- GT says incorrect but verifier says correct

This is the worst level.

## Required Output

Return only well-formed XML in exactly this structure:

<assessment>
Explain step-by-step:
1. whether GT treats the Proof Solution as correct or incorrect,
2. whether the verifier treats it as correct or incorrect,
3. which verifier-claimed errors match GT in step and reason,
4. whether the GT first fatal error was found,
5. which GT recoverable errors were matched or missed,
6. whether there are false positives,
7. why the final feedback level was assigned.

Mention step indices explicitly when discussing matched, missed, or false-positive errors.

<level>EXACT_MATCH|BASIC_MATCH|POOR_MATCH|WRONG_POLARITY</level>

## INPUT

## Problem Statement

{problem}

## Reference Solution (Optional)

{human_solution}

## Proof Solution

{solution}

## Solution Verification

{solution_verification}

## Human Ground Truth

{human_ground_truth}
\end{tcblisting}

\section{Annotation Protocol} \label{ap}

\subsection{Annotation Interface}

Each annotation instance presents the annotator with a mathematical problem, an optional official solution, a model-generated proof, and several auxiliary verifications, including model-generated assessments, error analyses, and first-error-step predictions. These traces are provided only for reference: annotators are instructed not to trust them by default, and the final label must be based on their own mathematical judgment of the proof.

\subsection{Annotation Principles}

Annotators are asked to judge the correctness of the full proof, rather than only the final conclusion. The following principles are applied throughout the annotation process:

\begin{itemize}
    \item \textbf{Mathematical validity.} Each step must be mathematically sound and must follow from previous steps, stated assumptions, or standard results whose conditions are satisfied.
    \item \textbf{Problem constraints.} If the problem imposes constraints on the method, domain, or form of the answer, the proof must respect these constraints.
    \item \textbf{Logical completeness.} A claim is accepted only when it is sufficiently justified in the proof. Plausible but unsupported assertions are treated conservatively and should be explicitly noted.
    \item \textbf{Whole-proof review.} Annotators should inspect the complete proof chain and record all relevant issues up to the first fatal error, rather than stopping at the first minor local defect.
\end{itemize}

\subsection{Error Categories}

To characterize the topology of proof failures, we distinguish two types of errors.

\begin{itemize}
    \item \textbf{Fatal Error.} A fatal error is a serious mathematical flaw that invalidates the global reasoning chain or prevents the proof from establishing the desired conclusion. Examples include an invalid theorem application, an unjustified implication that is essential to the argument, proving only one direction of an equivalence, or treating a necessary condition as sufficient.
    \item \textbf{Recoverable Error.} A recoverable error is a local issue that can be repaired by a minor correction or additional justification without changing the main proof strategy. Examples include notation slips, minor algebraic mistakes, small omitted justifications, or local sign-convention issues.
\end{itemize}

If an annotator is uncertain whether an issue should be treated as fatal, recoverable, or not erroneous, the case is flagged for further discussion and adjudication.

\subsection{Annotation Fields}

For each proof trajectory, annotators fill in the fields listed in Table~\ref{tab:annotation_fields}. Step indices always refer to the zero-based indices in the model-generated proof.

\begin{table}[htb]
    \centering
    \small
    \renewcommand{\arraystretch}{1.30}
    \begin{tabularx}{\columnwidth}{@{}p{0.28\columnwidth}X@{}}
        \toprule
        Field & Description \\
        \midrule
        Reviewer Comment & A concise but complete assessment of the proof. The comment should summarize which parts are correct, which parts fail, and why the proof is valid or invalid. Annotators are asked to evaluate the entire proof, not only the first erroneous step. \\
        Remark & Optional notes for abnormal cases, such as rendering errors, missing sections, mismatched problem-proof pairs, corrupted content, or flawed problem statements. \\
        First Fatal Error Step & The index of the first fatal error. The value is $-1$ if the proof contains no fatal error. \\
        Recoverable Error Steps & The indices of recoverable errors before the first fatal error. The value is $-1$ if no recoverable error is found. Multiple indices are separated by commas. Errors after the first fatal error are not required, since the global proof chain has already been invalidated. \\
        Report Error Case & A flag for abnormal or unjudgeable cases. When this flag is used, the annotator must explain the reason in the Remark field. \\
        \bottomrule
    \end{tabularx}
    \caption{Fields collected in the expert annotation protocol.}
    \label{tab:annotation_fields}
\end{table}
\FloatBarrier

\subsection{Label Semantics}

The expert label is determined by the combination of fatal and recoverable error annotations. A proof is treated as fully correct only when \textit{First Fatal Error Step} is $-1$ and \textit{Recoverable Error Steps} is also $-1$. If a fatal error is present, the proof is invalid regardless of later steps. Recoverable errors indicate local defects that should be reflected in the reviewer comment and meta-verification target, but they do not by themselves necessarily invalidate the global proof.

This protocol supports both coarse validity evaluation and fine-grained meta-verification. The first fatal error identifies the earliest step at which the proof becomes globally invalid, while recoverable error steps record local issues that a high-quality verifier should mention before the fatal cutoff. The reviewer comment provides the semantic explanation used to judge whether a model verifier has identified the correct mathematical issue, rather than merely predicting the correct error polarity.

\section{Training Details}
\label{app:training_details}

We employ Intern-S2-Preview-35B as the base verifier and gpt-oss-120b as the RL judge. The verifier is trained using the GRPO algorithm with asymmetric clipping ($\epsilon_{\text{low}}=0.2$, $\epsilon_{\text{high}}=0.28$, dual-clip $c=2.0$) and no KL penalty. During rollout, we use a global batch size of 128 and a prompt repetition factor of 16, with a maximum response length of 64K tokens and a sampling temperature of 1.0. The optimizer is AdamW with a constant learning rate of $5\times10^{-7}$, $\beta=(0.9, 0.95)$, and a weight decay of 0.1.
The RL judge performs meta-verification on each rollout trajectory and assigns scores of $1.0$, $0.75$, $0.5$, and $0$ to \texttt{EXACT\_MATCH}, \texttt{BASIC\_MATCH}, \texttt{POOR\_MATCH}, and \texttt{WRONG\_POLARITY}, respectively.

\section{Error Analysis}
\label{app:error_analysis}

In this section, we present our taxonomies of proof-generation and proof-verification failures.
We also present the detailed error distributions of GPT-5.5-xhigh and Intern-S2-Preview-35B on \textsc{ProverBench} and \textsc{VerifierBench}.

\subsection{Generation Error Taxonomy}
\label{app:prover_error_taxonomy}

Our generation error taxonomy contains four groups and ten error types spanning proof strategy,
assumptions, intermediate reasoning, and proof structure. The detailed breakdown, definitions, and examples are presented below.

\begin{tcolorbox}[taxonomygroup]
\textbf{Group A: Strategy-Level Errors}\\[-1pt]
\small The model selects an incorrect proof objective or misinterprets the
problem at the outset.
\end{tcolorbox}

\begin{tcolorbox}[taxonomyitem]
\textbf{A1. Wrong Proof Intent}

\smallskip
\noindent\textsc{Definition.}\enspace
The model misjudges the proposition's truth polarity or intended direction.

\smallskip
\noindent\textsc{Example.}\enspace
The model declares a true proposition false and attempts to construct a
counterexample, although the proposition follows from the implicit function
theorem.
\end{tcolorbox}

\begin{tcolorbox}[taxonomyitem]
\textbf{A2. Problem Misinterpretation}

\smallskip
\noindent\textsc{Definition.}\enspace
The model misreads a core mathematical object, the scope of a constraint, or
the surrounding context and consequently solves a different problem.

\smallskip
\noindent\textsc{Example.}\enspace
The model replaces the two-sided condition $|x|>M$ with $x>M$, thereby
omitting the case $x\to-\infty$.
\end{tcolorbox}

\begin{tcolorbox}[taxonomygroup]
\textbf{Group B: Assumption Errors}\\[-1pt]
\small The proof pursues the intended objective but relies on an unsupported
premise or applies a theorem outside its valid conditions.
\end{tcolorbox}

\begin{tcolorbox}[taxonomyitem]
\textbf{B1. Unjustified Assumption}

\smallskip
\noindent\textsc{Definition.}\enspace
The model introduces a condition absent from the problem, usually to simplify
the proof.

\smallskip
\noindent\textsc{Example.}\enspace
The model assumes that a subspace $F$ is finite-dimensional even though no
such condition is given.
\end{tcolorbox}

\begin{tcolorbox}[taxonomyitem]
\textbf{B2. Theorem Misapplication}

\smallskip
\noindent\textsc{Definition.}\enspace
The model invokes a theorem without verifying its prerequisites or while
violating them.

\smallskip
\noindent\textsc{Example.}\enspace
The model asserts uniform convergence of a series solution without
establishing the required compatibility conditions.
\end{tcolorbox}

\begin{tcolorbox}[taxonomygroup]
\textbf{Group C: Reasoning Errors}\\[-1pt]
\small The overall strategy is plausible, but an intermediate step contains
an unsupported inference or an incorrect mathematical operation.
\end{tcolorbox}

\begin{tcolorbox}[taxonomyitem]
\textbf{C1. Unjustified Logical Leap}

\smallskip
\noindent\textsc{Definition.}\enspace
The model asserts a nontrivial intermediate conclusion without adequate
proof, often through unsupported uses of ``clearly'' or ``it follows.''

\smallskip
\noindent\textsc{Example.}\enspace
The model assumes without proof that an invariant subspace decomposes as a
direct sum of intersections with eigenspaces.
\end{tcolorbox}

\begin{tcolorbox}[taxonomyitem]
\textbf{C2. Circular Reasoning}

\smallskip
\noindent\textsc{Definition.}\enspace
The target conclusion, or an equivalent proposition, is used as a premise in
its own derivation.

\smallskip
\noindent\textsc{Example.}\enspace
The model assumes that a relation is a partial order while proving its
defining properties, then invokes acyclicity to establish antisymmetry even
though the two claims are equivalent in that setting.
\end{tcolorbox}

\begin{tcolorbox}[taxonomyitem]
\textbf{C3. Mathematical Fact Error}

\smallskip
\noindent\textsc{Definition.}\enspace
An intermediate mathematical assertion is false, such as an incorrect
formula, inequality, algebraic property, or quantitative result.

\smallskip
\noindent\textsc{Example.}\enspace
The model calls $\alpha$ transcendental over $\mathbb{F}_3$ even though
$\alpha$ is a root of a polynomial in $\mathbb{F}_3[x]$ and is therefore
algebraic.
\end{tcolorbox}

\begin{tcolorbox}[taxonomyitem]
\textbf{C4. Computational Error}

\smallskip
\noindent\textsc{Definition.}\enspace
The proof framework is plausible, but a local algebraic, calculus, or counting
operation is performed incorrectly.

\smallskip
\noindent\textsc{Example.}\enspace
The model uses the wrong row when computing $P(Y=1)$, which propagates to
incorrect values for $E[Y]$, the covariance, and the correlation coefficient.
\end{tcolorbox}

\begin{tcolorbox}[taxonomygroup]
\textbf{Group D: Structural Errors}\\[-1pt]
\small These errors affect proof completeness or internal traceability rather
than the high-level proof direction.
\end{tcolorbox}

\begin{tcolorbox}[taxonomyitem]
\textbf{D1. Incomplete Argument}

\smallskip
\noindent\textsc{Definition.}\enspace
The proof omits a necessary case, boundary condition, or key derivation, or
terminates before establishing the claimed conclusion.

\smallskip
\noindent\textsc{Example.}\enspace
The proof uses limits at $z\pm\delta$ without ensuring that these points are
continuity points of the relevant distribution function.
\end{tcolorbox}

\begin{tcolorbox}[taxonomyitem]
\textbf{D2. Citation Error}

\smallskip
\noindent\textsc{Definition.}\enspace
The proof cites an incorrect internal step or lemma even though the underlying
mathematical statement is otherwise correct.

\smallskip
\noindent\textsc{Example.}\enspace
Step 5 attributes an equality to Step 4 even though it was derived in
Steps 2--3.
\end{tcolorbox}

\subsection{Prover Error Distribution}
\label{app:prover_error_distribution}

Table~\ref{tab:prover_error_distribution_full} reports the detailed
distribution of proof-generation errors for GPT-5.5-xhigh and Intern-S2-Preview-35B on \textsc{ProverBench}.
Entries report counts, with percentages of all failed generations shown in parentheses.

\begin{table}[htb]
    \centering
    \begin{minipage}{0.78\linewidth}
    \centering
    \normalsize
    \captionsetup{font=small}
    \setlength{\tabcolsep}{6pt}
    \renewcommand{\arraystretch}{1.1}
    \begin{tabularx}{\linewidth}{@{}>{\raggedright\arraybackslash}X*{2}{>{\raggedleft\arraybackslash}p{0.22\linewidth}}@{}}
        \toprule
        Error Type & \shortstack{GPT-5.5-\\xhigh} & \shortstack{Intern-S2-\\Preview-35B} \\
        \midrule
        \textbf{Strategy-Level Errors} & \textbf{41 (43.6\%)} & \textbf{48 (27.1\%)} \\
        \quad Wrong Proof Intent & 27 (28.7\%) & 36 (20.3\%) \\
        \quad Problem Misinterpretation & 14 (14.9\%) & 12 (6.8\%) \\
        \midrule
        \textbf{Assumption Errors} & \textbf{12 (12.8\%)} & \textbf{17 (9.6\%)} \\
        \quad Unjustified Assumption & 10 (10.6\%) & 16 (9.0\%) \\
        \quad Theorem Misapplication & 2 (2.1\%) & 1 (0.6\%) \\
        \midrule
        \textbf{Reasoning Errors} & \textbf{26 (27.7\%)} & \textbf{81 (45.8\%)} \\
        \quad Unjustified Logical Leap & 13 (13.8\%) & 41 (23.2\%) \\
        \quad Circular Reasoning & 4 (4.3\%) & 11 (6.2\%) \\
        \quad Mathematical Fact Error & 8 (8.5\%) & 18 (10.2\%) \\
        \quad Computational Error & 1 (1.1\%) & 11 (6.2\%) \\
        \midrule
        \textbf{Structural Errors} & \textbf{2 (2.1\%)} & \textbf{18 (10.2\%)} \\
        \quad Incomplete Argument & 1 (1.1\%) & 10 (5.6\%) \\
        \quad Citation Error & 1 (1.1\%) & 8 (4.5\%) \\
        \midrule
        Other & 13 (13.8\%) & 13 (7.3\%) \\
        \midrule
        Total failures & 94 & 177 \\
        \bottomrule
    \end{tabularx}
    \caption{Detailed distribution of proof errors on \textsc{ProverBench}.}
    \label{tab:prover_error_distribution_full}
    \end{minipage}
\end{table}
\FloatBarrier

\subsection{Proof Verification Error Analysis}
\label{app:verifier_error_taxonomy}

\subsubsection{False-Negative Taxonomy}

We distinguish two complementary verification failures. A false negative
rejects an expert-verified proof, whereas a false positive accepts a proof
containing at least one expert-annotated error.
False negatives are organized into the following two verifier-side failure
types. For false positives, we instead classify the underlying missed defect
using the error taxonomy in Section~\ref{app:prover_error_taxonomy}.

\begin{tcolorbox}[taxonomyitem]
\textbf{Over-strict Penalization}

\smallskip
\noindent\textsc{Definition.}\enspace
The verifier rejects a valid proof for omitting conventional intermediate
steps, using an equivalent boundary convention, or containing an
inconsequential presentational defect.

\smallskip
\noindent\textsc{Example.}\enspace
The verifier rejects a valid stability argument because it uses a standard
linearization approximation without restating the routine intermediate
derivation.
\end{tcolorbox}

\begin{tcolorbox}[taxonomyitem]
\textbf{Interpretation Mismatch}

\smallskip
\noindent\textsc{Definition.}\enspace
The verifier adopts one interpretation of a mathematical object, domain, or
convention and rejects an equivalent formulation used by the proof.

\smallskip
\noindent\textsc{Example.}\enspace
The verifier expects open intervals to generate the Borel
$\sigma$-algebra and rejects a valid proof using closed intervals, although
the two generating families are equivalent.
\end{tcolorbox}

\subsection{Verifier Error Distribution}
\label{app:verifier_error_distribution}

Table~\ref{tab:verifier_error_distribution_full} reports the detailed
false-negative and false-positive distributions for GPT-5.5-xhigh and
Intern-S2-Preview-35B on \textsc{VerifierBench}. Percentages are computed separately within each
model's total number of FNs or FPs, and the subtypes under each outcome describe
the composition of that outcome.

\medskip
{\centering
\begin{minipage}{0.78\linewidth}
    \centering
    \normalsize
    \captionsetup{font=small}
    \setlength{\tabcolsep}{6pt}
    \renewcommand{\arraystretch}{1.1}
    \begin{tabularx}{\linewidth}{@{}>{\raggedright\arraybackslash}X*{2}{>{\raggedleft\arraybackslash}p{0.22\linewidth}}@{}}
        \toprule
        Error Type & \shortstack{GPT-5.5-\\xhigh} & \shortstack{Intern-S2-\\Preview-35B} \\
        \midrule
        \textbf{False Negatives} & 95 (100.0\%) & 17 (100.0\%) \\
        \quad Over-strict Penalization & 94 (98.9\%) & 12 (70.6\%) \\
        \quad Interpretation Mismatch & 0 (0.0\%) & 5 (29.4\%) \\
        \quad Other & 1 (1.1\%) & 0 (0.0\%) \\
        \midrule
        \textbf{False Positives} & 112 (100.0\%) & 251 (100.0\%) \\
        \quad \textbf{Strategy-Level Errors} & \textbf{11 (9.8\%)} & \textbf{34 (13.5\%)} \\
        \qquad Wrong Proof Intent & 0 (0.0\%) & 1 (0.4\%) \\
        \qquad Problem Misinterpretation & 11 (9.8\%) & 33 (13.1\%) \\
        \quad \textbf{Assumption Errors} & \textbf{1 (0.9\%)} & \textbf{5 (2.0\%)} \\
        \qquad Unjustified Assumption & 0 (0.0\%) & 4 (1.6\%) \\
        \qquad Theorem Misapplication & 1 (0.9\%) & 1 (0.4\%) \\
        \quad \textbf{Reasoning Errors} & \textbf{74 (66.1\%)} & \textbf{152 (60.6\%)} \\
        \qquad Unjustified Logical Leap & 51 (45.5\%) & 100 (39.8\%) \\
        \qquad Circular Reasoning & 3 (2.7\%) & 6 (2.4\%) \\
        \qquad Mathematical Fact Error & 5 (4.5\%) & 14 (5.6\%) \\
        \qquad Computational Error & 15 (13.4\%) & 32 (12.7\%) \\
        \quad \textbf{Structural Errors} & \textbf{25 (22.3\%)} & \textbf{59 (23.5\%)} \\
        \qquad Incomplete Argument & 14 (12.5\%) & 45 (17.9\%) \\
        \qquad Citation Error & 11 (9.8\%) & 14 (5.6\%) \\
        \quad Other & 1 (0.9\%) & 1 (0.4\%) \\
        \bottomrule
    \end{tabularx}
    \captionof{table}{Detailed verification error distribution on \textsc{VerifierBench}.}
    \label{tab:verifier_error_distribution_full}
\end{minipage}
\par}

\end{document}